\documentclass[]{dataflow}

\usepackage[toc,page,header]{appendix}

\usepackage{minitoc}
\usepackage{multirow}     
\usepackage{multicol}     
\usepackage{booktabs}     
\usepackage{colortbl}     
\usepackage[table]{xcolor} 
\usepackage{makecell}     
\usepackage{graphicx}     
\usepackage{minted}
\usepackage{xcolor}
\usepackage{fontawesome5}

\title{DataFlow: An LLM-Driven Framework for Unified Data Preparation and Workflow Automation in the Era of Data-Centric AI}

\author[*, \dagger]{Hao Liang}
\author[*, \dagger]{Xiaochen Ma}
\author[*,\dagger]{Zhou Liu}
\author[*]{Zhen Hao Wong}
\author[*]{Zhengyang Zhao}
\author[*]{Zimo Meng}
\author[*]{Runming He}
\author[*]{Chengyu Shen}
\author[*]{Qifeng Cai}
\author[*]{Zhaoyang Han}
\author[*]{Meiyi Qiang}
\author[*]{Yalin Feng}
\author[*]{Tianyi Bai}
\author[]{Zewei Pan}
\author[]{Ziyi Guo}
\author[]{Yizhen Jiang}
\author[]{Jingwen Deng}
\author[]{Qijie You}
\author[]{Peichao Lai}
\author[]{Tianyu Guo}
\author[]{Chi Hsu Tsai}
\author[]{Hengyi Feng}
\author[]{Rui Hu}
\author[]{Wenkai Yu}
\author[]{Junbo Niu}
\author[]{Bohan Zeng}
\author[]{Ruichuan An}
\author[]{Lu Ma}
\author[]{Jihao Huang}
\author[]{Yaowei Zheng}
\author[]{Conghui He}
\author[]{Linpeng Tang}
\author[]{Bin Cui}
\author[]{Weinan E}
\author[\ddagger]{Wentao Zhang}

\affiliation[]{$^{1}$Peking University, $^{2}$Institute for Advanced Algorithms Research, Shanghai, $^{3}$OriginHub Technology, $^{4}$OpenDataLab, Shanghai Artificial Intelligence Laboratory, $^{5}$LLaMA-Factory Team}

\abstract{

The rapidly growing demand for high-quality data in Large Language Models (LLMs) has intensified the need for scalable, reliable, and semantically rich data preparation pipelines. However, current practices remain dominated by ad-hoc scripts and loosely specified workflows, which lack principled abstractions, hinder reproducibility, and offer limited support for model-in-the-loop data generation. To address these challenges, we present DataFlow, a unified and extensible LLM-driven data preparation framework. DataFlow is designed with system-level abstractions that enable modular, reusable, and composable data transformations, and provides a PyTorch-style pipeline construction API for building debuggable and optimizable dataflows. The framework consists of nearly 200 reusable operators and six domain-general pipelines spanning text, mathematical reasoning, code, Text-to-SQL, agentic RAG, and large-scale knowledge extraction. To further improve usability, we introduce DataFlow-Agent, which automatically translates natural-language specifications into executable pipelines via operator synthesis, pipeline planning, and iterative verification. Across six representative use cases, DataFlow consistently improves downstream LLM performance. Our math, code, and text pipelines outperform curated human datasets and specialized synthetic baselines, achieving up to +3\% execution accuracy in Text-to-SQL over SynSQL, +7\% average improvements on code benchmarks, and 1–3 point gains on MATH, GSM8K, and AIME. Moreover, a unified 10K-sample dataset produced by DataFlow enables base models to surpass counterparts trained on 1M Infinity-Instruct data. These results demonstrate that DataFlow provides a practical and high-performance substrate for reliable, reproducible, and scalable LLM data preparation, and establishes a system-level foundation for future data-centric AI development.

}

\date{\today}

\def\emailicon{\raisebox{-1.5pt}{\includegraphics[height=1.05em]{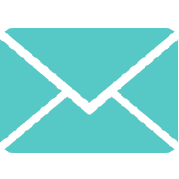}}}
\def\githubicon{\raisebox{-1.5pt}{\includegraphics[height=1.05em]{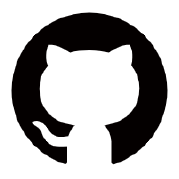}}}
\def\huggingfaceicon{\raisebox{-1.5pt}{\includegraphics[height=1.05em]{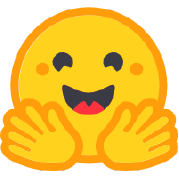}}}

\contribution[*]{Equal Contribution}
\contribution[\dagger]{Project Leader}
\contribution[\ddagger]{Corresponding author}

\checkdata[ \emailicon \hspace{0.3em} Correspondence ]{\email{wentao.zhang@pku.edu.cn}}

\newcommand{\sourcelink}{https://github.com/OpenDCAI/DataFlow}
\newcommand{\datalink}{https://huggingface.co/datasets/OpenDCAI/dataflow-instruct-10k}

\checkdata[ \githubicon \hspace{0.3em} Source Code ]{ \url{\sourcelink} }
\checkdata[ \huggingfaceicon \hspace{0.3em} Dataset ]{ \url{\datalink} }

\checkdata[ \faFile \hspace{0.57em} Codebase Documentation ]{ \url{https://opendcai.github.io/DataFlow-Doc/} }

\begin{document}
\maketitle

\renewcommand{\thefootnote}{\fnsymbol{footnote}} 
\setcounter{footnote}{0}

\renewcommand{\thefootnote}{\arabic{footnote}}
\pagestyle{fancy}
\fancyhf{}
\fancyhead[L]{DataFlow Technical Report}
\fancyhead[R]{\thepage}

\newpage
\tableofcontents
\newpage

\section{Introduction}


Large language models (LLMs) have rapidly evolved from research prototypes to foundational infrastructure across natural language processing and beyond. Since OpenAI introduced the GPT~\cite{achiam2023gpt} family through large-scale human annotation and ignited the era of large language models (LLMs), scaling-law studies~\cite{hoffmann2022training, sorscher2022beyond} have consistently demonstrated that data quality and quantity are central to model performance.
As model scales continue to grow and downstream tasks become increasingly complex, the size and semantic diversity of training corpora have expanded dramatically~\cite{hui2024qwen2, yang2025qwen3}. Modern LLM development now relies on multi-stage, semantics-heavy data preparation pipelines that integrate synthetic, refinement, filtering, and domain-specific transformation across trillions of tokens~\cite{wang2025ultra, chen2024data, nvidia2024nemo}.


However, despite the critical role of high-quality data, data preparation for LLMs remains fragmented and largely unstandardized. Most practitioners still rely on ad-hoc scripts and loosely standardized workflows, which lack explicit dataflow abstractions, well-defined atomic operators, or any form of pipeline-level optimization.
The absence of a unified and programmable paradigm makes pipelines difficult to reproduce, extend, or compare across projects~\cite{chen2024data, nvidia2024nemo, ostendorff2024llm}.
This problem is amplified by the trend toward increasingly fine-grained post-training tasks, such as instruction tuning, chain-of-thought generation, or function calling, where both the semantic richness and the semantic accuracy in data preparation are essential for achieving precise task-level model behavior~\cite{zhang2023instruction, wei2022chain}.


In response to this fragmentation, several systems have recently emerged with the goal of standardizing LLM data curation. Frameworks such as NeMo Curator~\cite{nvidia2024nemo} and Data-Juicer~\cite{chen2024data} offer substantial functionality—including captioning, rewriting, classification, and multimodal processing—and have significantly improved the efficiency of large-scale corpus construction. Yet these systems remain fundamentally extraction- and filtering-oriented, and their abstractions provide limited support for expressing iterative, model-in-the-loop generative workflows with fine-grained semantic control. As a result, they are ill-suited for pipelines in which data synthesis and multi-step semantic refinement are central rather than auxiliary.

This limitation is becoming increasingly consequential. LLMs are no longer only consumers of data, but also producers. Because large-scale human annotation is prohibitively expensive, recent work heavily leverages LLM-based data synthesis workflows to construct high-quality corpora at scale~\cite{bai2024survey}. Multiple recent reports show that, in many regimes, carefully synthesized data can outperform even high-quality selected data~\cite{wang2023self, yu2025cot}, further underscoring the importance of LLM-driven generation workflows.



Given these trends, we argue that a unified framework for LLM data preparation must elevate LLM-driven data synthesis to a first-class, programmable dataflow abstraction. Such a framework should: (1) provide fine-grained, composable operators for model-in-the-loop generation and semantic refinement; (2) support explicit, verifiable pipeline definitions that serve as an inspectable, domain-agnostic open-source protocol for LLM data preparation—much like how \verb|torch.nn.Module| standardizes model composition in deep learning; (3) remain backend-agnostic to integrate different LLM engines and storage backends; and (4) enable principled workflow composition, reuse, and optimization across models, tasks, and domains, while further supporting agent-driven automatic workflow construction. Taken together, these requirements signal a shift in data preparation—from post-hoc corpus cleaning toward LLM-centric workflows that build high-fidelity, semantically rich, and task-aligned synthetic corpora through iterative synthesis and refinement.

Motivated by this shift, we introduce \textsc{DataFlow}, a unified and automated LLM-driven framework for end-to-end LLM data preparation, with a \textsc{DataFlow-Agent} that allows users to compose pipelines directly from natural-language specifications. \textsc{DataFlow} places LLMs at the center of the operator ecosystem: most operators are LLM-driven, with a small number implemented using heuristics or small models. The framework provides over 180 operators organized into four categories—generation, evaluation,  filtering, and refinement—and includes more than 90 reusable prompt templates that enable operator-level composition and consistent behavior across tasks. Using these primitives, \textsc{DataFlow} includes a set of state-of-the-art (SOTA) synthesis pipelines that span mathematical reasoning, raw text, code, Text-to-SQL, agentic RAG-style data, and large-scale QA extraction from web or PDF corpora. All pipelines are expressed within \textsc{DataFlow}’s common abstractions and require no task-specific glue code, adhering to a generate–evaluate–filter workflow augmented with targeted refinement stages. 

To ensure usability, extensibility, and long-term maintainability, \textsc{DataFlow} adopts a PyTorch-like programming interface that exposes its core abstractions, global storage, LLM serving, operators, prompt templates, and pipelines, through modular Python classes and functions. 
This code-first design avoids complex YAML or shell-based configuration schemes and provides an IDE-friendly development workflow, including code completion and reliable navigation. 
Beyond the core library, operators, prompt templates, and pipelines can be developed outside the main repository and packaged as standalone Python modules, enabling practitioners to publish and reuse domain-specific components as first-class \textsc{DataFlow-Extensions}. 
To support this ecosystem, \textsc{DataFlow} includes a Command-Line Interface (CLI) toolchain that scaffolds new extension packages, from operator stubs to full pipeline repositories, standardizing development practices and lowering the barrier to community contribution. 
Finally, \textsc{DataFlow-Agent} serves as an agentic orchestration layer that translates natural-language specifications into executable pipelines and can automatically synthesize and debug new operators when needed, further accelerating the construction of scalable and semantically rich LLM-driven data preparation workflows.



Extensive experiments on six \textsc{DataFlow}-implemented pipelines show that our design philosophy is effective across diverse data preparation scenarios, consistently producing high-quality training data. Across all settings, the resulting \textsc{DataFlow} datasets match or even surpass SOTA baselines, including curated human datasets, specialized synthetic workflows, and the strong Qwen2.5-Instruct series. For example, \textsc{DataFlow}-synthesized mathematical reasoning data yields 1–3 point gains over high-quality synthetic baselines~\cite{openr1,2025synthetic1} on MATH, GSM8K, and AIME; our Text-to-SQL pipelines achieve over +3\% execution-accuracy improvements compared with the 2.5M-sample SynSQL corpus~\cite{li2025omnisql} while using less than 0.1M training examples; and \textsc{DataFlow}-based code pipelines deliver over 7\% average improvements relative to widely used public code instruction datasets~\cite{codealpaca,wei2024selfcodealign}.  

Moreover, by combining \textsc{DataFlow}-generated text, math, and code data into a unified corpus, \textsc{DataFlow-Instruct-10K}, we find that training on only 10K samples enables Qwen2-base and Qwen2.5-base to surpass models trained on 1M Infinity-Instruct~\cite{li2025infinity} instances, while approaching the performance of their corresponding Qwen-Instruct model. This demonstrates that \textsc{DataFlow} can produce domain-diverse supervision of sufficiently high quality to yield substantial gains in data efficiency.

Together, these results demonstrate that \textsc{DataFlow} is not only an end-to-end system for LLM-based data preparation, but also a comprehensive operator and algorithm library and an open, user-friendly protocol framework. Built around six SOTA template pipelines and a large collection of reusable operators, \textsc{DataFlow} offers a unified foundation for LLM-centric data construction, enabling principled, semantically rich, and scalable workflows that improve programmability, reproducibility, and data quality across domains.

Overall, our key contributions are summarized as follows:

\begin{itemize}

\item \textbf{A unified LLM-driven data preparation framework.}
We propose \textsc{DataFlow}, a unified system for LLM data preparation built on composable abstractions and an LLM-first operator execution model.

\item \textbf{A rich and extensible operator–pipeline ecosystem.}
\textsc{DataFlow} provides nearly 200 reusable operators and six SOTA template pipelines covering text, mathematical reasoning, code, Text-to-SQL, agentic RAG data, and large-scale QA extraction.

\item \textbf{A developer- and open-source–friendly programming model.}
Through a PyTorch-like API, IDE-native tooling, and plugin-style extensibility via Python packages, \textsc{DataFlow} enables reproducible experimentation, easy customization, and community-driven extensions to form \textsc{DataFlow-Ecosystem}.

\item \textbf{An agentic orchestration layer for automated pipeline construction.}
\textsc{DataFlow-Agent} composes executable pipelines from natural-language intent, lowering the barrier to building scalable and semantically rich LLM-driven workflows.

\item \textbf{Extensive empirical validation and open-source data release.}
Experiments across six pipelines show that \textsc{DataFlow}-generated data consistently improves downstream LLM performance and data efficiency. We additionally release a high-quality, multi-domain dataset produced entirely with \textsc{DataFlow} to support further research and benchmarking.

\end{itemize}
\section{Background and Related Works}
\subsection{Data in LLM Development}
The development of LLMs involves several key stages, among which training is particularly crucial, as the model learns fundamental linguistic patterns from large-scale corpora. During this stage, the model is exposed to vast amounts of text data from various domains, enabling it to acquire a broad understanding of language. 

Consequently, the quality and diversity of training data directly impact the model’s ability to generalize effectively across different contexts~\cite{li2024datacomp, gao2020pile}. Recently, the rapid development of large language models has brought about a substantial increase in the volume of training data~\cite{llama, achiam2023gpt}. In this scenario, the quality and quantity of data become even more paramount. 

High-quality data can significantly enhance model performance~\cite{llama3repo}. As the volume of data increases, ensuring high data quality becomes more challenging, as it requires additional resources for data cleaning, selection, and annotation~\cite{bai2024survey}. Poor-quality data can cause models to learn incorrect patterns and produce inaccurate predictions. Furthermore, insufficient data diversity may result in models performing well in specific domains but exhibiting poor generalization in cross-domain tasks. Additionally, distributional shifts in the data can exacerbate model over-reliance on training distributions, diminishing their applicability in real-world scenarios.

\subsection{Data Preparation for LLMs}

As disclosed by the above discussion, data preparation is a crucial step in training LLMs, significantly impacting the model's performance and generalization capabilities. With the continuous expansion of LLM scales, the complexity and efficiency of data preparation have become key research focuses. However, although systems like Apache Spark~\cite{Spark}, Dask~\cite{rocklin2015dask}, and Hadoop~\cite{ghemawat2003google, dean2008mapreduce, white2012hadoop} are powerful for large-scale Extract–Transform–Load (ETL), they are not a good fit for modern LLM data preparation. These frameworks can, in principle, run semantic cleaning by calling LLMs or embedding models as user-defined functions, but they provide no native support for model-in-the-loop processing, GPU-efficient batching, or token-level text operations. More importantly, their built-in operators focus on structured data and offer very limited functionality for unstructured text, meaning that essential steps—such as tokenization, language detection, document segmentation, semantic deduplication, or safety filtering—must be implemented manually with ad-hoc User-Defined Functions (UDFs). This leads to significant overhead and engineering complexity, making general big-data engines inadequate for the large-scale, semantics-heavy pipelines required for LLM corpus construction.

LLM-based methods have been widely used in data quality evaluation and data selection. For instance, MoDS~\cite{du2023mods} leverages DeBERTa for scoring and retaining high-quality data, while Alphagasus~\cite{chen2023alpagasus} uses ChatGPT to score data accuracy. Other studies have employed GPT-4 for data rewriting and quality improvement. For a comprehensive overview, refer to the data for LLM survey~\cite{bai2024survey}.

\subsection{Existing LLM Data Preparation Systems}\label{sec:existing_data_preparation_system}
Recent work increasingly approaches LLM training data preparation as a first-class systems problem. 
Table~\ref{table:data-prep-compare} summarizes the distinguishing characteristics of the major frameworks.

\textbf{NeMo Curator}~\cite{nvidia2024nemo}  is an open-source, GPU-accelerated library from NVIDIA that offers modular pipelines for large-scale LLM data curation, including data download and extraction (e.g., Common Crawl, arXiv, Wikipedia), language identification, text cleaning, heuristic and learned quality filtering, domain and toxicity classification, document- and semantic-level deduplication, privacy filtering, and even synthetic data generation, all built on Dask/RAPIDS and designed to scale to multi-node, multi-GPU environments.

\textbf{Data-Juicer}~\cite{chen2024data} is a ``one-stop'' data processing system that abstracts LLM data recipes into composable operators: the original system already provides 50+ operators for constructing and evaluating text data mixtures, while the 2.0 version extends this to 100+ operators across text, image, video, and audio, supporting analysis, cleaning, synthesis, annotation, and post-training data pipelines with tight integration to Ray and HuggingFace Datasets.

These systems substantially improve the efficiency and quality of LLM data preparation, but they remain largely configuration-centric toolkits. In contrast, our framework is built around a rich library of nearly 200 reusable text-specific operators, enabling fine-grained control over cleaning, transformation, synthesis, and evaluation; multiple pipelines instantiated from these operators consistently yield strong downstream gains, and even simple mixtures of data produced by different pipelines remain highly effective. Moreover, the system adopts a modular, PyTorch-style ``building-block'' design with lightweight, well-defined interfaces, making it natural for data agents to compose, orchestrate, and invoke data-processing pipelines programmatically.

\begin{table}[t]
\centering
\caption{High-level comparison of existing data preparation systems for LLM.}
\resizebox{\linewidth}{!}{
\begin{tabular}{lccc}
\toprule
\textbf{Dimension} &
\textbf{Data-Juicer~\cite{chen2024data}} &
\textbf{NeMo Curator~\cite{nvidia2024nemo}} &
\textbf{\textsc{DataFlow} (ours)} \\
\midrule
\textbf{Primary focus} &
Filtering / Cleaning &
Large-scale Curation &
LLM-driven Synthesis + Refinement \\
\textbf{Programming model} &
Config-based Recipes &
Component-based Pipelines &
PyTorch-like Operators \& Pipelines \\
\textbf{LLM integration} &
Partial (some gen ops) &
Minimal (mainly filtering) &
First-class Serving + Templates \\
\textbf{Automation} &
Recommendation Agent &
None &
Pipeline Construct/Debug Agent \\
\textbf{Extensibility} &
OperatorZoo / Cookbook &
Custom Scripts &
Extension Packages + CLI Scaffolding \\
\bottomrule
\end{tabular}
}
\label{table:data-prep-compare}
\end{table}

\section{DataFlow System Overview}
\label{sec:df_overview}
In this section, we present a overview of \textsc{DataFlow} a unified and automated system that standardizes and streamlines multi-domain data preparation for LLMs.

\subsection{Goals and Design Philosophy}
DataFlow is designed around six core goals:
\paragraph{Ease of Use.} 
A PyTorch-inspired~\cite{paszke2019pytorch}, IDE-friendly programming interface enables users to build and debug complex data preparation pipelines with minimal boilerplate.

\paragraph{Extensibility.}
Following a modular abstraction similar to \texttt{torch.nn.Module}, new operators, and algorithms can be added as plug-and-play components and naturally compose with existing workflows.

\paragraph{Unified Paradigm.}
\textsc{DataFlow} unifies heterogeneous data preparation workflows under a standardized abstraction layer. The design balances \emph{standardization}, for consistency and reproducibility, with \emph{customization} needed across domains, enabling efficient pipeline reuse and adaptation.

\paragraph{Performance Efficiency.} The official pipelines in \textsc{DataFlow} achieve performance comparable to or exceeding SOTA data preparation methods, demonstrating that unification does not impose substantial overhead.

\paragraph{Intelligent Automation.} A lightweight agentic subsystem leverages core abstractions to interpret natural-language intent and automatically construct or adjust operators and pipelines, supporting rapid prototyping and reducing manual engineering.

\paragraph{Open Source Paradigm.} 
\textsc{DataFlow} aims to serve as a community standard for LLM data preparation. Its unified abstractions enable reproducible pipeline sharing, transparent swapping of LLM backends, and controlled experimentation.





\subsection{System Scope and Positioning}
\begin{figure}
    \centering
    \includegraphics[width=\linewidth]{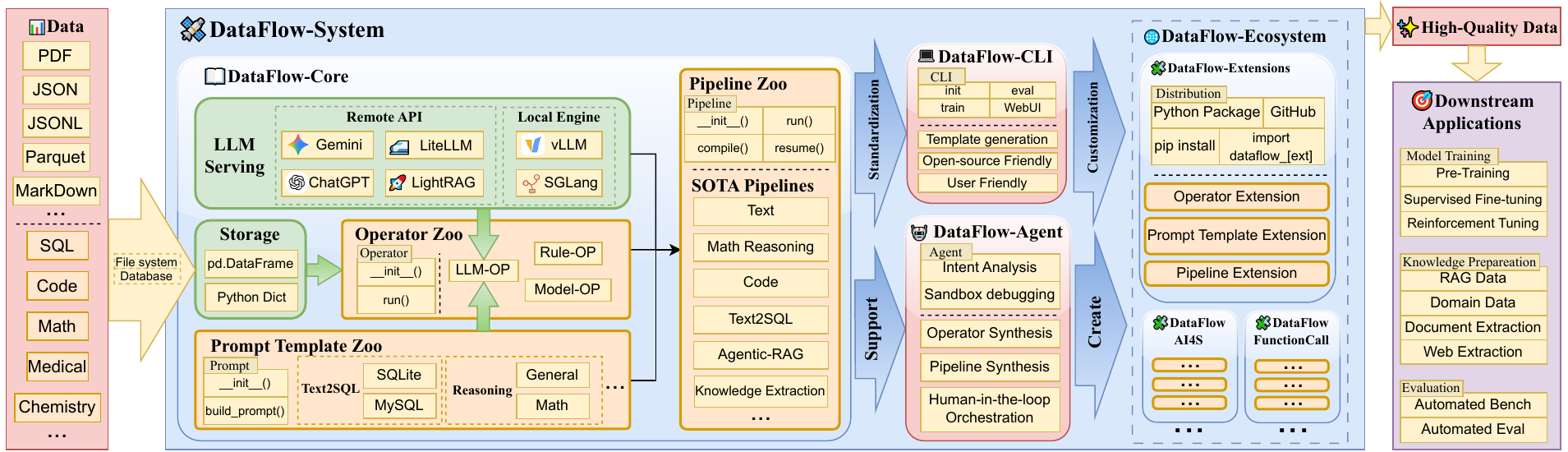}
    \caption{
        High-level architecture of \textsc{DataFlow}. 
        The system consists of a core execution engine (storage, operators, templates, and LLM serving), 
        reusable pipelines, user-facing control layers (CLI and agent), and an extensible ecosystem for domain-specialized workflows. 
        \textsc{DataFlow} produces high-quality, task-aligned datasets consumed by downstream LLM applications.
    }
    \label{fig:df_overview}
\end{figure}

\textsc{DataFlow} spans the full workflow of LLM-centric data preparation. As Figure \ref{fig:df_overview} shown, at its core, the system provides unified abstractions for storage, LLM serving, operators, prompt templates, and pipelines—defining the execution substrate on which all transformations are performed.  
Above the core, two user-facing control layers, the CLI and the \textsc{DataFlow-Agent}, support both scriptable and automated workflow construction.  
Beyond the engine, \textsc{DataFlow-Extensions} offer a modular interface for adding Python-package–based operators, templates, and pipelines. Domain-specialized packages built on this interface collectively form the broader \textsc{DataFlow-Ecosystem}.  
Together, these components define the system boundary: \textsc{DataFlow} provides the abstractions and control layers for data preparation, while downstream LLM training, evaluation, and retrieval applications consume its outputs.

\subsection{System Workflow}

Figure~\ref{fig:df_overview} also illustrates the end-to-end workflow of \textsc{DataFlow}. The system ingests datasets from common file formats (e.g., JSON, JSONL, CSV, Parquet, Markdown, PDF) as well as domain-specific sources such as SQL logs and code repositories, converting all inputs into a unified tabular representation maintained by the core storage layer. Operators interact with this shared storage to read and write intermediate results, enabling consistent data flow across transformation stages.

Operators implement transformations such as generation, refinement, filtering, and evaluation. LLM-driven operators invoke local inference engines (e.g., vLLM~\cite{kwon2023efficient_vllm}, SGLang~\cite{zheng2024sglang}) or online API-based services (e.g., Gemini~\cite{team2024gemini}, ChatGPT~\cite{achiam2023gpt}) via the unified serving abstraction, while rule-based and small-model operators execute independently of LLM backends.

Pipelines in the Pipeline Zoo compose these operators into reusable workflows for tasks such as text synthesis, mathematical reasoning, code processing, Text-to-SQL generation, agentic RAG, and large-scale knowledge extraction. Pipelines may be executed directly, compiled for optimized execution, resumed from intermediate states, or adapted to new domains.

Users interact with \textsc{DataFlow} during workflow execution through either the CLI or the \textsc{DataFlow-Agent}: the CLI issues explicit execution commands, while the agent translates natural-language specifications into executable workflows and performs iterative debugging. Workflow outputs, high-quality, task-aligned datasets, integrate seamlessly into downstream LLM applications.

\section{Framework Design and Architecture}
This section presents the internal design of \textsc{DataFlow} and formalizes the execution model underlying its abstractions in Section~\ref{sec:df_overview}. 
\textsc{DataFlow} is organized around four architectural pillars: 
(1) a global storage abstraction that maintains the canonical tabular representation of datasets and mediates all data access; 
(2) a set of hierarchical programming interfaces for LLM serving, operators, prompt templates, and pipelines; 
(3) a principled operator categorization scheme that reconciles open-ended domain requirements with a compact set of reusable transformation primitives; and 
(4) an extension mechanism that supports a growing ecosystem of user-contributed components. 
Together, these elements provide a scalable and extensible substrate for constructing, executing, and sharing LLM-centric data preparation workflows.

\subsection{Global Storage Abstraction and Operator Interaction}
At the core of \textsc{DataFlow}'s execution substrate is a unified storage abstraction that maintains the canonical tabular representation of the dataset and mediates all data access during workflow execution. 
LLM-oriented data—such as instructions, responses, chain-of-thought traces, scores, and metadata—is naturally expressed as key–value fields associated with each sample, making a tabular structure a suitable and expressive organizational format. 
The storage layer decouples data management from operator logic, exposing a minimal, backend-agnostic API through the \texttt{DataFlowStorage} base class. 
This design allows custom storage backends—such as file-system, object-store, or database implementations—to be integrated without altering operator behavior.

The abstraction provides two primary operations:
\begin{itemize}
    \item \texttt{read()}: retrieve the current dataset (or relevant fields) in a format required by the operator.
    \item \texttt{write(data)}: update or append fields to the shared dataset representation.
\end{itemize}

\begin{figure}[ht]
    \centering
    \includegraphics[width=0.8\linewidth]{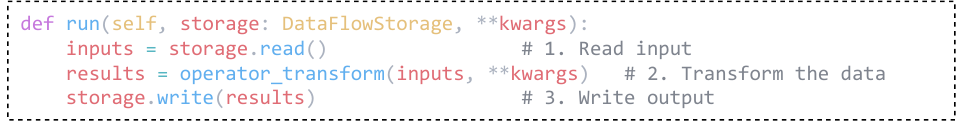}
    \caption{The standard execution pattern of an operator’s \texttt{run()} method in \textsc{DataFlow}. 
Within \texttt{run()}, the operator interacts with the global \texttt{DataFlowStorage} by retrieving inputs through \texttt{storage.read()}, applying its transformation logic, and writing updated fields back via \texttt{storage.write()}. 
This read--transform--write paradigm captures how data flows from one operator to the next throughout the workflow.}
    \label{fig:operator_storage}
\end{figure}

Centralizing all access through these operations ensures that operators remain agnostic to physical storage layout, while intermediate artifacts produced by one operator become immediately available to others. 
A typical operator interaction follows the pattern in Figure~\ref{fig:operator_storage}.

Because operators operate only against this logical abstraction, they can be reordered, recomposed, or batched without modifying their internals, and improvements to the storage backend (e.g., adding distributed or database-backed implementations) require no operator-level changes. 
The default storage implementation uses a Pandas as the execution substrate and supports common input/output formats such as \texttt{JSON}, \texttt{JSONL}, \texttt{CSV}, and \texttt{Parquet}.


\subsection {Hierarchical Programming Interfaces}

\textsc{DataFlow} exposes a hierarchical programming interface built around four core abstractions. 
(1) The serving interface provides a unified mechanism for issuing LLM inference requests across heterogeneous backends. 
(2) Operators define reusable data-transformation units and may optionally invoke the serving layer when LLM-driven computation is required. 
(3) Prompt templates specify how operator inputs are rendered into concrete prompts and how model outputs should be structured or constrained, providing a declarative interface for consistent prompt construction. 
(4) Pipelines compose operators into multi-stage workflows with explicit data dependencies and support optional compilation for validation and optimization. 
The following subsections describe these abstractions in detail.



\subsubsection{LLM Serving API}

LLM-driven operators rely on a unified serving API that abstracts over heterogeneous model backends. 
The API exposes a single high-level entry point, 
\texttt{generate\_from\_input(user\_inputs, system\_prompt, json\_schema)}, 
which accepts a list of prompts, typically assembled by the calling operator, and returns a list of model-generated outputs. 
Optional arguments such as a \texttt{system\_prompt} or an output \texttt{json\_schema} enable structured prompting and decoding when needed. 
This interface shields operators from backend-specific considerations such as batching, retry strategies, request routing, and rate limiting.

The serving layer supports both:
\begin{itemize}
    \item \emph{Local inference engines} (e.g., vLLM~\cite{kwon2023efficient_vllm}, SGLang~\cite{zheng2024sglang}), which exploit backend-level parallelism for high-throughput execution; and
    \item \emph{Online API-based services} (e.g., ChatGPT~\cite{achiam2023gpt}, Gemini~\cite{team2024gemini}), for which \textsc{DataFlow} performs multi-threaded request dispatch to maximize throughput.
\end{itemize}

This unified serving abstraction reduces the implementation burden of LLM-driven operators and enables flexible backend substitution, making it easy to assess how different LLM choices influence data preparation quality.

\begin{figure}
    \centering
    \includegraphics[width=1.0\linewidth]{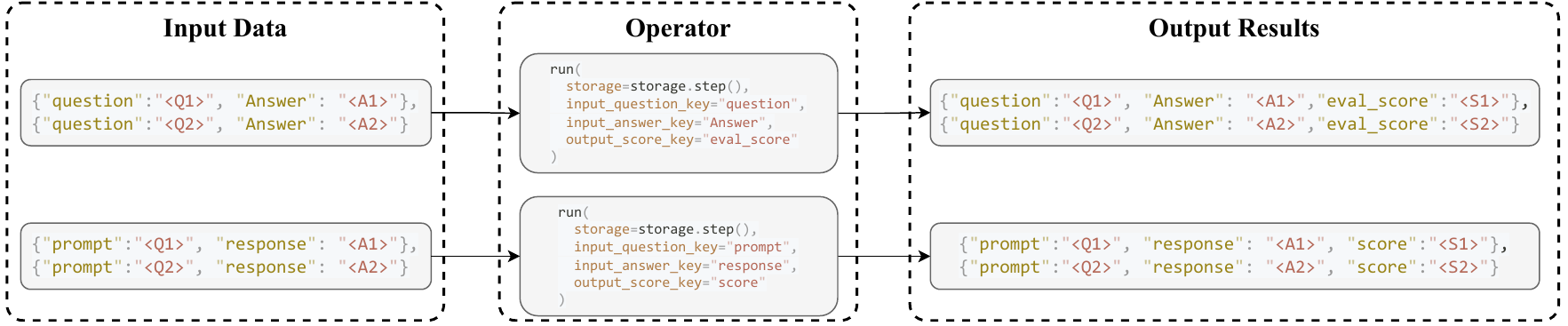}
\caption{
Example of how an operator’s \texttt{run()} method interacts with data via key-based bindings.
This flexible key-binding mechanism adapts to arbitrary datasets without preprocessing and enables seamless operator composition.
}
    \label{fig:run_func_for_op}
\end{figure}
\subsubsection{Operator Programming Interface}

Operators serve as the fundamental transformation units in \textsc{DataFlow}. 
They follow a two-phase interface that cleanly separates initialization from execution: initialization configures the operator, while execution performs the transformation. 
This separation allows heterogeneous behaviors, from LLM-driven generation to rule-based filtering, to be expressed under a unified abstraction.

During initialization (\texttt{\_\_init\_\_()}), an operator receives configuration parameters such as hyperparameters or task-specific settings. 
LLM-driven operators may additionally bind to a LLM serving object and a prompt-template object in this stage, whereas rule-based and lightweight-model operators omit these bindings entirely. 
Initialization therefore captures all static configuration and external dependencies, leaving execution to focus exclusively on data transformation.

An operator’s \texttt{run()} method implements its transformation logic and constitutes the unit of execution within a pipeline. 
To keep operators general and easily composable, \texttt{run()} accepts only a \texttt{DataFlowStorage} object together with a set of \texttt{input\_*} and \texttt{output\_*} keys. 
Interpreting these as key–value pairs, an \texttt{input\_*} key indicates the storage column to be read as an input field, while an \texttt{output\_*} key indicates the name of the new column to be written for each processed data item. 
Figure~\ref{fig:run_func_for_op} illustrates this mapping. 
This design provides flexible I/O bindings that naturally adapt to diverse upstream datasets, while the declared keys form a directed dependency graph among operators, enabling topological scheduling and downstream optimization checks.

By isolating configuration from execution and constraining state changes to explicit key-based read/write operations on shared storage, the operator abstraction remains lightweight, deterministic, and easy to compose. 
These properties allow \textsc{DataFlow} to support a wide range of transformation behaviors under a single, portable interface while preserving consistent execution semantics throughout the system.

\subsubsection{Prompt Template Interface}
Prompts serve as the primary mechanism guiding LLMs to perform task-specific transformations. 
Every LLM-driven operator relies on a prompt, and operators that share the same high-level logic often differ only in subtle prompt variations. 
For instance, in Text-to-SQL generation, synthesizing queries for SQLite and MySQL involves identical operator logic; the only difference lies in minor syntax adjustments communicated through the prompt. 
To support such reuse while accommodating domain-specific variations, \textsc{DataFlow} decouples prompt construction from operator implementation through a dedicated prompt template interface.

A prompt template encapsulates a reusable prompt pattern and provides parameterized slots that operators populate at execution time. 
Each LLM-driven operator initializes its associated template during \texttt{\_\_init\_\_()}, following the same configuration–execution paradigm as other system components. 
During execution, the operator invokes the template’s \texttt{build\_prompt()} method, which assembles task-relevant information—such as input fields, schema hints, or contextual metadata—into a concrete prompt that is subsequently passed to the LLM serving layer. 
This encapsulation allows the operator’s transformation logic to remain agnostic to how prompts are rendered.

To facilitate one-to-many mappings between operators and templates, LLM-driven operators expose a unified \texttt{op.ALLOWED\_PROMPTS} interface that enumerates all compatible prompt templates. 
This design enables operators to be flexibly reused across domains or tasks by simply switching or tuning templates, without modifying operator logic.

Overall, the prompt template interface provides a declarative mechanism for prompt construction, promotes operator reuse across closely related tasks, and ensures consistent prompting behavior throughout \textsc{DataFlow}'s LLM-driven workflows.

\begin{figure}[t]
    \centering
    \includegraphics[width=1.0\linewidth]{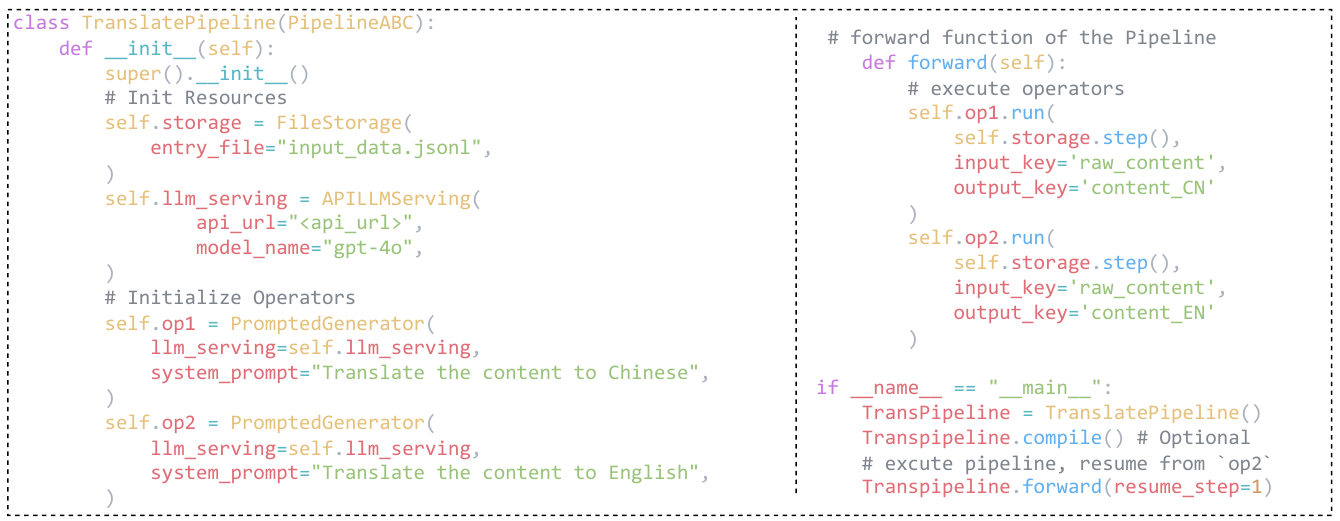}
    \caption{Illustration of the \textsc{DataFlow} pipeline API. 
The example shows how a pipeline declares its storage and serving backends, instantiates operators with task-specific configurations, and executes them via \texttt{forward()} using input/output key bindings. 
The interface supports compilation and stepwise resumption, enabling flexible and modular workflow construction.}
    \label{fig:pipeline_api}
\end{figure}

\subsubsection{Pipeline Composition Interface}

Building on the abstractions introduced above, \textsc{DataFlow} provides a pipeline interface that enables users to compose operators into multi-stage data-preparation workflows. 
A pipeline is represented as an ordered sequence of operators (or a lightweight DAG ), forming an end-to-end execution graph that captures the intended dataflow. 
Figure~\ref{fig:pipeline_api} illustrates the pipeline API and its core components.

The pipeline API adopts a PyTorch~\cite{paszke2019pytorch}-like design in which the \texttt{\_\_init\_\_()} method handles resource allocation and operator configuration, while the \texttt{()} method encodes a single pass of execution. 
Within \texttt{forward()}, operator-specific key bindings implicitly define the dataflow topology, allowing pipelines to be constructed in a modular, readable, and IDE-friendly manner.

Functionally, the pipeline interface provides a built-in \texttt{compile()} procedure that performs static analysis of the operator sequence prior to execution. 
During compilation, \textsc{DataFlow} extracts operator dependencies and parameters, constructs the corresponding DAG, and conducts key-level validation to detect missing fields, type inconsistencies, and malformed dependency chains. 
Instead of executing operators immediately, \texttt{compile()} records all operator configurations and dependency information to produce a deferred execution plan. 
This deferred-construction design follows the Factory Method pattern~\cite{gamma1995design}, in which object creation is separated from object execution: the actual invocation of each operator’s \texttt{run()} method is deferred until the subsequent \texttt{forward()} call.

The compiled execution graph first provides complete structural information to the \textsc{DataFlow}-Agent, enabling it to surface all key- and dependency-related errors in a single report. 
This significantly reduces the number of debugging rounds required by the agent and lowers the associated inference cost. 
Additionally, the compiled graph defines a minimal and efficient execution plan that supports advanced runtime features such as checkpointing and stepwise resumption, improving iterative development and large-scale pipeline construction.

\subsection{Operator Categorization}

Operators in \textsc{DataFlow} encapsulate diverse data-processing algorithms that, when composed, support end-to-end LLM data preparation workflows. 
As a unified yet extensible framework intended to serve arbitrarily many domains, \textsc{DataFlow} must simultaneously accommodate an open-ended set of domain-specific algorithms while exposing a stable and comprehensible operator space. 
These competing forces—unbounded domain requirements and the need for conceptual compactness—introduce inherent tension. 
To reconcile this, \textsc{DataFlow} organizes operators along multiple orthogonal categorization dimensions. 
Categories are mutually exclusive within each dimension, while dimensions themselves are parallel. 
This categorization scheme has been validated across the diverse domains covered in this paper, including more than six state-of-the-art data preparation pipelines, demonstrating both its representational sufficiency and scalable generality.

\paragraph{Modality Dimension.}
The fundamental categorization separates operators by the modality they process, such as text, visual content, or document-like inputs. 
Modalities must be distinguished because operators within the same modality share compatible input–output semantics and can interoperate, whereas operators across different modalities often cannot be composed directly. 
\textsc{DataFlow} primarily operates on textual representations, with non-text modalities first processed by modality-specific operators that parse or convert raw inputs—such as images or PDFs—into text before any downstream transformations are applied. 
Therefore, Clear modality classification makes this conversion flow explicit and enables the pipeline compiler to validate operator chains, ensuring that modality transitions are correctly specified and that only compatible operators are composed.

\paragraph{Core vs.\ Domain-Specific Dimension.}
A second categorization distinguishes between \emph{core operators} and \emph{domain operators}. 
Core operators reflect the fundamental design philosophy of \textsc{DataFlow} and serve as the conceptual basis from which most other operators can be derived. 
Although domain operators may wrap or specialize core operators, their semantics can generally be expressed by instantiating the parameters of a corresponding core operator. 
Core operators are intentionally limited in number and relatively stable, forming the recommended entry point for new users. 
Domain operators, by contrast, expand without bound as new domains, modalities, or tasks emerge. 
Although theoretically unbounded, the domain operators included in \textsc{DataFlow} are limited to those required to support the best-performing pipelines across existing domains, ensuring practical conciseness and avoiding unnecessary proliferation.

\begin{figure}
    \centering
    \includegraphics[width=1.0\linewidth]{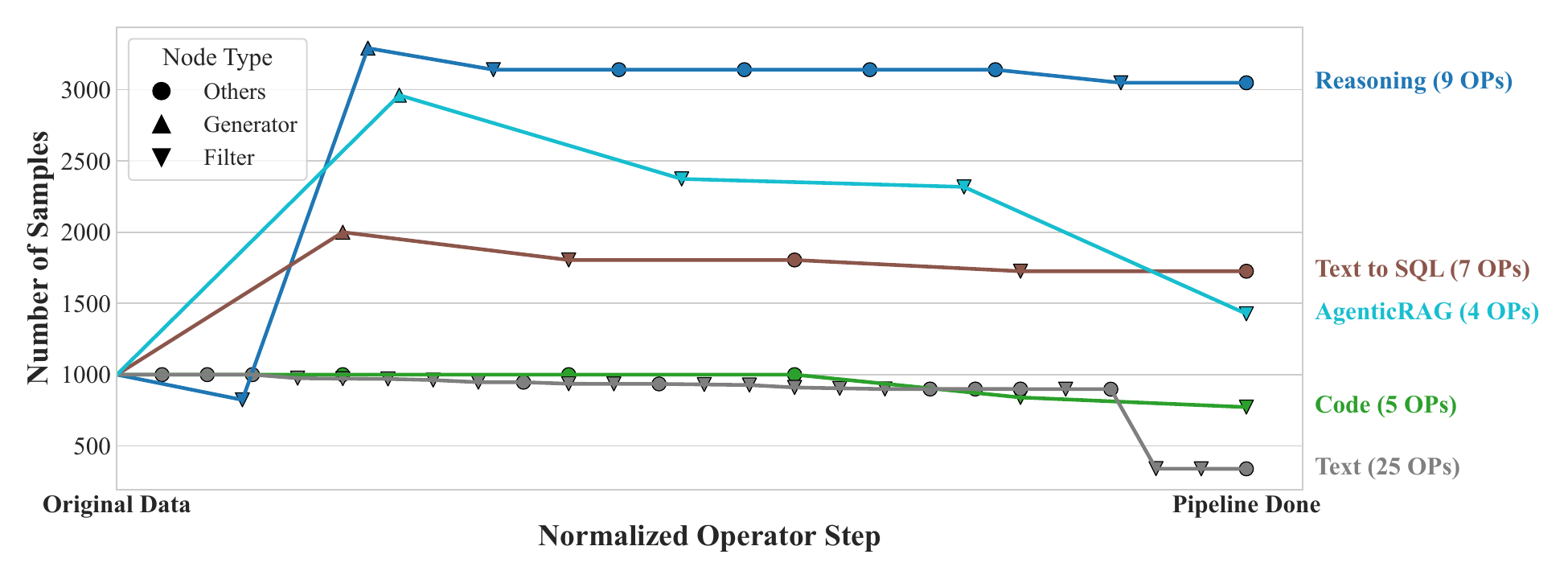}
\caption{
Evolution of sample counts across operator stages in \textsc{DataFlow} pipelines.
All pipelines start with 1000 input samples.
The Text pipeline mainly performs pre-training data filtering, and the Code pipeline focuses on expanding code capabilities based on existing instruction data; therefore, neither of these pipelines involves any generative components.
}

    \label{fig:op_trend}
\end{figure}

\paragraph{Functional Dimension.}
At a finer granularity, operators fall into four functional categories—\emph{generate}, \emph{evaluate}, \emph{filter}, and \emph{refine}—each capturing a distinct transformation pattern in data preparation. 
These categories align with a core design philosophy of \textsc{DataFlow} as a data-synthesis framework: pipelines first expand the candidate space through generation, then score and filter the results, optionally applying refinement stages in between. 
This generate–evaluate–filter–refine paradigm underlies most pipeline designs in \textsc{DataFlow}. 
As illustrated in Figure~\ref{fig:op_trend}, when a pipeline begins with 1{,}000 input samples, the number of data items typically increases during generation stages and then contracts as evaluation, filtering, and refinement operators are applied.

To make this paradigm concrete, \textsc{DataFlow} defines four operator categories, each with clear semantics and naming conventions. 
Throughout this discussion, we use the tabular representation adopted by \textsc{DataFlow}: each row denotes a data sample, and each field corresponds to a named column within that sample.

\begin{itemize}
    \item \textbf{Generate.}  
        These operators augment data by adding new textual fields or producing additional rows.  
        Operators ending with \texttt{Generator} add new fields to existing rows, whereas those ending with \texttt{RowGenerator} increase the number of rows.  
        Example usages include generating answers to questions.

    \item \textbf{Evaluate.}  
        These operators compute scores or labels for either individual samples or entire datasets.  
        \texttt{SampleEvaluator} operators attach evaluation metadata to each row, whereas \texttt{DatasetEvaluator} operators output dataset-level metrics.  
        Examples include assigning difficulty levels to math problems or classifying QA pairs by subject.

    \item \textbf{Filter.}  
        These operators reduce the number of rows based on criteria derived from existing fields or evaluation results.  
        Their semantics maintain row contents apart from newly added evaluation fields.  
        Examples include removing samples with incorrect answers.

    \item \textbf{Refine.}  
        These operators modify specific fields within existing rows without changing the number of samples.  
        They often apply lightweight transformations such as removing URLs or emojis from text.  
        Operators typically end with the suffix \texttt{Refiner}.
\end{itemize}

Across these dimensions, \textsc{DataFlow} supports both systematic extensibility and bounded conceptual complexity: 
the modality and core-versus-domain dimensions organize an open-ended operator ecosystem, while the functional dimension provides a compact and reusable set of transformation primitives for constructing scalable LLM data preparation workflows.

\subsection{DataFlow-Ecosystem}

A unified data preparation framework must accommodate an open-ended set of algorithms and workflows, which naturally leads to an unbounded space of operators and pipelines. 
To structure this extensibility in a maintainable manner, \textsc{DataFlow} introduces the concept of a \emph{\textsc{DataFlow}-Extension}: a modular package that encapsulates additional operators, prompt templates, and pipelines. 
User-contributed extensions collectively form the broader \emph{\textsc{DataFlow}-Ecosystem}, a plug-and-play environment analogous to Python’s package ecosystem, where practitioners can readily publish, share, and reuse domain-specific components.

To streamline extension development, \textsc{DataFlow} provides automated project scaffolding through the \textsc{DataFlow-CLI}. 
Given a few high-level specifications, the CLI generates ready-to-use templates for operators, prompt templates, pipelines, and even full repository layouts suitable for distribution via PyPI or GitHub. 
Developers need only implement task-specific logic within these generated stubs. 
Both the core system and extension packages can be installed and imported through Python’s package manager, while lazy-loading mechanisms ensure that multiple extensions coexist with minimal environmental interference.

Complementing the CLI, the \textsc{DataFlow-Agent} supports natural-language–driven construction of operators and pipelines. 
Leveraging the domain knowledge embedded in large language models, the agent synthesizes effective data-transformation logic and automates common design steps, substantially reducing the cost of authoring high-quality \textsc{DataFlow}-Extensions.

Together, the \textsc{DataFlow-CLI} and \textsc{DataFlow-Agent} reduce the overhead of extension development and promote community-driven growth. 
Our goal is to cultivate a sustainable open-source ecosystem in which data preparation recipes—constructed from standardized operators, prompt templates, and pipelines—can be shared, reproduced, and improved, ultimately accelerating progress across the data-centric ML community.


\section{DataFlow-Agent}
\label{sec:df_agent}
The \textsc{DataFlow-Agent} serves as the intelligent orchestration layer atop the \textsc{DataFlow} framework. 
It bridges high-level human intent with low-level data-processing execution by leveraging the modular abstractions of DataFlow together with a graph-based multi-agent workflow engine.  
Built on LangGraph~\cite{langgraph2024}, the agent layer coordinates a set of specialized agents through a stateful execution graph, translating natural-language directives into executable, self-correcting, and optimized data preparation pipelines.

\begin{figure}[t]
    \centering
    \includegraphics[width=1\linewidth]{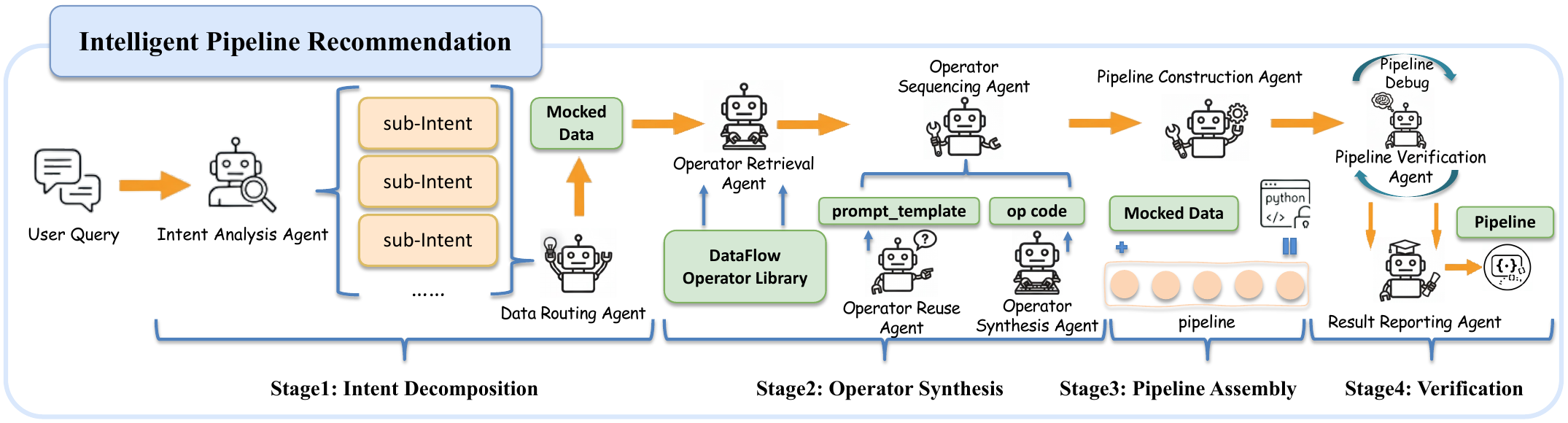}
    \caption{\textsc{DataFlow-Agent} architecture: a LangGraph-orchestrated multi-agent workflow that translates natural-language intent into a verified executable DAG pipeline.}
    \label{fig:dfa-pdf}
\end{figure}

\subsection{AgentRoles}
To achieve autonomous pipeline construction and code synthesis, the system decomposes responsibilities across a roster of specialized agents. Each agent encapsulates specific logic and interacts with the \textsc{DataFlow} core components:

\begin{itemize}
    \item \textbf{Intent Analysis Agent:} Accepts the user's high-level natural language query and decomposes it into a structured sequence of actionable sub-intents, providing the foundational blueprint for the pipeline.
    
    \item \textbf{Data Routing Agent:} Analyzes the provided input data to determine the task category for routing, or generates synthetic data placeholders if no data is supplied to enable dry-run execution.
    
    \item \textbf{Operator Retrieval Agent:} Takes specific sub-intents as input and employs RAG to retrieve the most relevant existing operators from the \textsc{DataFlow} library as potential candidates.
    
    \item \textbf{Operator Sequencing Agent:} Evaluates candidate operators for I/O compatibility to select the best fit, or outputs detailed specifications for new operators when functional gaps are detected.
    
    \item \textbf{Operator Synthesis Agent:} Receives specifications for missing functions and generates context-aware code using RAG, performing automated unit-level debugging until the code is executable.
    
    \item \textbf{Operator Reuse Agent:} Assesses the generated operator code for quality and creates a reusable \texttt{prompt\_template}, ensuring the code can be efficiently reused without rewriting.
    
    \item \textbf{Pipeline Construction Agent:} Orchestrates the assembly of all validated operators (both pre-existing and newly synthesized) into a coherent Directed Acyclic Graph (DAG) structure ready for processing.
    
    \item \textbf{Pipeline Verification Agent:} Executes the assembled pipeline within a sandboxed environment to identify runtime errors, autonomously adjusting connections or parameters to output a validated, error-free pipeline.
    
    \item \textbf{Result Reporting Agent:} Synthesizes the final workflow details and execution results, generating a comprehensive report and an executable pipeline artifact as the final solution.
\end{itemize}

\subsection{Intelligent Pipeline Recommendation}
As shown in Figure \ref{fig:dfa-pdf}, the core capabilities of the system are realized through a sophisticated agentic layer built atop the DataFlow framework. This layer employs LangGraph \cite{langgraph2024} to orchestrate a series of specialized agents within graph-based stateful workflows. 

\paragraph{Intent Decomposition}
The workflow begins when the system receives a user's natural language query. The Intent Analysis Agent decomposes this high-level objective into a sequence of discrete, actionable sub-intents. Concurrently, the Data Routing Agent evaluates the input dataset to categorize the task for downstream routing. If no dataset is provided, this agent generates synthetic data placeholders to enable a complete dry-run execution.

\paragraph{Operator Synthesis}
To fulfill these sub-intents, the Operator Retrieval Agent searches the \textsc{DataFlow} library for relevant operators, which the Operator Sequencing Agent evaluates for compatibility. If a functional gap is identified, the Operator Reuse Agent first assesses whether the requirement can be met by reusing existing code via a \texttt{prompt\_template}. Only when reuse is not feasible does the Operator Synthesis Agent generate new code using RAG-based few-shot learning. The code is then debugged automatically to ensure stable execution.

\paragraph{Pipeline Assembly}
After all retrieved or synthesized operators are validated, the Pipeline Construction Agent assembles them into a single pipeline. It represents the pipeline as a DAG and defines the initial connections so data can flow from the source to the sink.

\paragraph{Verification}
The system then runs an integration test. The Pipeline Verification Agent executes the pipeline in a sandbox with a data sample to check connectivity and runtime behavior. If errors occur, it fixes them by adjusting parameters or connections. After the pipeline passes validation, the Result Reporting Agent generates a report and outputs the final executable pipeline definition.

\subsection{Summary}
In summary, unlike Data-Juicer's agentic approach \cite{chen2024data}, which is largely constrained to parameterizing and sequencing a static library of pre-existing operators, \textsc{DataFlow-Agent} achieves a significantly higher degree of autonomy through its ability to dynamically synthesize and debug executable code for missing functionalities. By integrating a "retrieve-reuse-synthesize" strategy with a self-correcting verification loop, our system transcends simple configuration generation, enabling the construction of truly adaptive pipelines that can handle unforeseen requirements without manual coding intervention.

\section{Use Cases \& Pipelines}

\textsc{DataFlow} integrates a rich collection of data pipelines covering diverse text centric task domains, including text processing, mathematical reasoning data, Text-to-SQL generation, and agentic data preparation. In addition, \textsc{DataFlow} supports structured knowledge extraction and normalization from PDFs and textbooks, enabling tasks such as schema construction, domain grounding, and instruction synthesis.

All pipelines are implemented through reusable operators and declarative workflow specifications, allowing users to flexibly compose, extend, and adapt them to new scenarios with minimal engineering effort. More detailed tutorials, pipeline examples, and operator-level documentation are available at the website: \href{https://opendcai.github.io/DataFlow-Doc/}{https://opendcai.github.io/DataFlow-Doc/}.

\begin{figure*}[t] 
\centering
\includegraphics[width=1.0\textwidth]{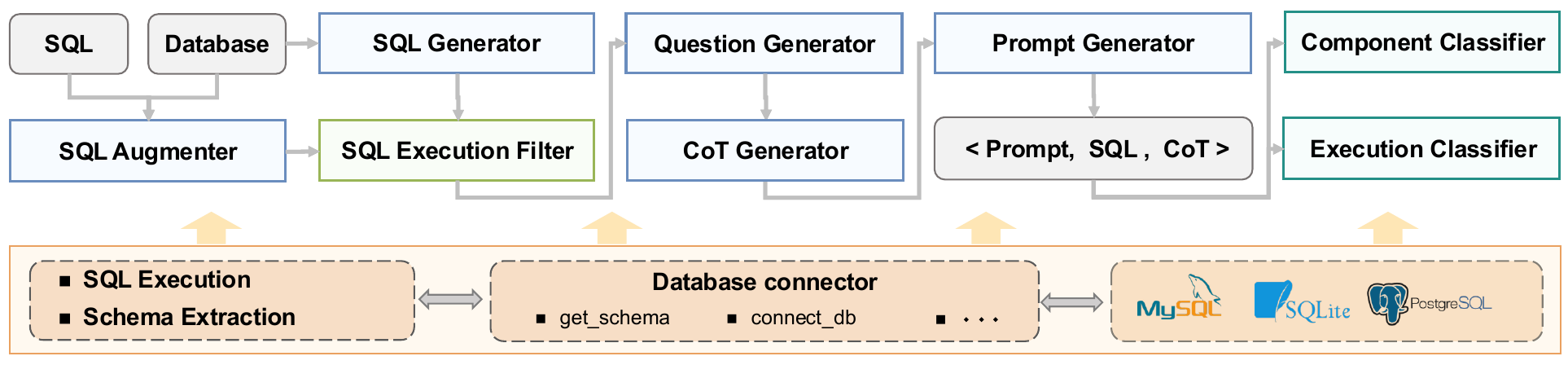}
\caption{Overall framework of Text-to-SQL pipelines in \textsc{DataFlow}.}
\label{fig:text2sql_pipeline}
\end{figure*}

\subsection{Case Study: Text-to-SQL Data Pipeline in \textsc{DataFlow}}
We first design a set of specifically designed, reusable Text-to-SQL operators to ensure modularity and extensibility (see Section \ref{sec:text2sql_operators}). As shown in Figure~\ref{fig:text2sql_pipeline}, we introduce two pipelines to construct high-quality Text-to-SQL datasets (see Section \ref{sec:text2sql_pipelines}). Furthermore, Section \ref{sec:text2sql_supports} describes the support for database operations and the prompt template mechanisms provided by \textsc{DataFlow}.

\subsubsection{Operators}
\label{sec:text2sql_operators}

\paragraph{SQL Generator.}
The SQL Generator operator produces SQL queries from scratch using the database, ensuring both diversity and validity. Four levels of complexity, simple, moderate, complex, and highly complex, are defined and randomly selected to guide the LLM in generating queries of varying difficulty through clear definitions and few-shot examples. The database schema, including \texttt{CREATE TABLE} statements for all relational tables and randomly sampled column values, provides the necessary context for the LLM to understand the database. Advanced SQL functions are also randomly supplied to increase the realism of the generated queries. Since natural language questions often require querying specific entries, the number of returned columns is constrained accordingly. Under task instructions, the LLM produces meaningful SQL queries. Within the \textsc{DataFlow} framework, the SQL Generator operator can be naturally adapted and reused across different databases (e.g., MySQL, SQLite, PostgreSQL) simply by replacing the corresponding prompt template.

\paragraph{SQL Augmentor.}
The SQL Augmentor operator generates diverse, closely related augmented SQL queries based on seed SQL rather than synthesizing them from scratch. We propose six augmentation strategies to expand SQL queries in different directions: (1) Data Value Transformation, (2) Query Structure Modification, (3) Business Logic Alteration, (4) Complexity Enhancement, (5) Introduction of Advanced SQL Features, and (6) Performance and Optimization. Categories are randomly selected and applied through few-shot prompting. The database schema and values are provided as contextual information. Given an original SQL query and task instructions, the augmentor produces its augmented SQL counterpart.

\paragraph{Text2SQL Consistency Filter.}
For existing pairs of natural language questions and SQL queries, inconsistencies may arise where the two do not correspond. Such problematic data needs to be filtered out. This is achieved using an LLM, which analyzes whether the question and SQL align in content.

\paragraph{SQL Execution Filter.}
Not all generated SQL queries are valid or efficient. Therefore, the SQL Execution Filter operator filters queries from two perspectives: (1) whether the SQL query can be successfully executed on the target database, and (2) whether its runtime exceeds a preset threshold, in which case it is discarded to ensure system responsiveness.

\paragraph{Question Generator.}
The Question Generator operator generates a semantically equivalent natural language question based on the SQL. Natural language questions are categorized into the following stylistic types: (1) Tone and Formality: formal vs. colloquial, (2) Syntactic Structure and Intent: imperative, interrogative, and declarative, (3) Information Density and Clarity: concise, descriptive, ambiguous, and metaphorical, and (4) Interaction Mode: role-playing and procedural. The first two categories cover queries with clear user intent, whereas ambiguous and metaphorical styles involve unclear or figurative language. A target language style is randomly selected, and the database schema is provided for context. Based on task instructions and the generated SQL query, the LLM produces a natural language question.

\paragraph{Chain-of-Thought Generator.}
Chain-of-Thought(CoT) reasoning enhances a model’s ability to solve complex tasks by breaking them down into a series of smaller, manageable sub-problems. To generate CoT reasoning traces, the task instructions, database schema, the generated natural language question, and the generated SQL query are needed. The LLM produces a complete reasoning chain covering intermediate reasoning steps and the final SQL query. During CoT validation, the generated SQL is extracted from the reasoning chain. A CoT process is considered a valid solution only if the execution result of its generated SQL matches that of the reference SQL on the given database.

\paragraph{Prompt Generator.}
As the primary input to the model, a prompt contains the necessary information for reasoning. To facilitate reliable Text-to-SQL generation, a well-structured prompt should include not only the natural language question but also the database schema and specific task instructions to guide the model. The Prompt Generation operator synthesizes these components into a final prompt.

\paragraph{SQL Component Classifier.}
Classifying SQL queries enables deeper analysis of their structural complexity. Following the evaluation standards of Spider~\cite{yu2018spider}, SQL queries are categorized into four difficulty levels, simple, moderate, hard, and extra hard, based on the number and complexity of their syntactic components. These components include column selections, the use of aggregate functions in the SELECT clause, and advanced constructs such as \texttt{GROUP BY}, \texttt{ORDER BY}, \texttt{INTERSECT}, or nested subqueries. The SQL Component Classifier operator assigns each SQL query to one of these categories according to the defined criteria.

\paragraph{SQL Execution Classifier.}
 Whether the model can generate correct SQL for a given natural language question is also a meaningful measure of difficulty. In the SQL Execution Classifier operator, LLM is instructed to generate SQL query $k$ times on the same input prompt and count the number of successful executions, denoted as $n$. We then classify the difficulty level based on $\frac{n}{k}$. Unlike the SQL component classifier operator, execution difficulty is model-dependent: more capable LLMs achieve higher success rates on the same task and thus are considered to have lower execution difficulty.

\subsubsection{Pipelines}
\label{sec:text2sql_pipelines}

In the design philosophy of \textsc{DataFlow}, pipelines are decomposed into independent operator units according to their functionalities, enabling maximal reusability of operators. As shown in Figure~\ref{fig:text2sql_pipeline}, the designed operators are composed into two pipelines to support SQL data synthesis in different scenarios.

\paragraph{SQL Generation Pipeline.}

This pipeline generates SQL from scratch based on the database schema. It first uses the SQL Generator operator to produce initial SQL statements, followed by the SQL Execution Filter to remove low-quality or non-executable SQL. Next, the Question Generator produces the natural language question corresponding to each SQL query, the Chain-of-Thought Generator operator generates the reasoning steps (CoT), and the Prompt Generator constructs the prompt content. Finally, the SQL Component Classifier and SQL Execution Classifier assign difficulty labels to the data.

\paragraph{SQL Refinement Pipeline.}

This pipeline generates data starting from the existing seed SQLs. The pipeline first verifies the quality of the seed SQL using the SQL Execution Filter, and the Text2SQL Consistency Filter removes samples where the SQL does not align with the natural language question. Then, the SQL Augmentor produces augmented SQL based on the seed SQL. The subsequent steps mirror those in the SQL Generation Pipeline: filtering low-quality SQL with the SQL Execution Filter, generating natural language questions via the Question Generator, producing CoT reasoning via the Chain-of-Thought Generator, composing prompts with the Prompt Generator, and finally assigning difficulty labels using the SQL Component Classifier and SQL Execution Classifier.

\subsubsection{\textsc{DataFlow}-support Mechanism}
\label{sec:text2sql_supports}

\paragraph{Database Manager Module.}
Within the Pipeline, an efficient and reliable data interaction mechanism serves as the core infrastructure that ensures the stable execution of the workflow. To this end, we implement the \texttt{Database Manager} module, which encapsulates the low-level details of database interaction and provides a unified, efficient, and extensible programming interface. The \texttt{Database Manager} improves processing throughput under high-concurrency workloads and abstracts schema metadata retrieval, thereby reducing the upper layers’ dependency on the underlying database structure.
To achieve cross-database compatibility, we introduce the abstract base class \texttt{DatabaseConnector}. This class defines a standardized set of interfaces, including \texttt{connect\_db} (establishing a database connection), \texttt{execute\_sql} (executing SQL statements and returning results), and \texttt{get\_schema} (retrieving complete schema metadata). For each database system, developers need only subclass this base class and implement the system-specific driver invocation and error-handling logic, enabling seamless integration into the overall system.

\paragraph{Prompt Template Module.}
When generating SQL, different scenarios, such as CRUD queries, vector search SQL, or SQL categorized by different difficulty specifications, require distinct prompt templates. To maximize operator reusability under these varying requirements, \textsc{DataFlow} introduces the Prompt Template module. This design allows the SQL Generator operator to be reused across scenarios by simply substituting the Prompt class. In practice, one only needs to reimplement the \texttt{build\_prompt} method within a new Prompt class, without modifying the SQL Generator operator itself.

\section{Experiments}
\begin{table}[t]
\centering
\small
\caption{Pre-training Data Filtering: Performance comparison across models trained with 30B-scale tokens on general evaluation benchmarks.}
\label{tab:PT}
\begin{tabular}{lccccccc}
\toprule
\textbf{Methods} & \textbf{ARC-C} & \textbf{ARC-E} & \textbf{MMLU} & \textbf{HellaSwag} & \textbf{WinoGrande} & \textbf{Gaokao-MathQA} & \textbf{Avg} \\
\midrule
\textbf{Random-30B}      & 25.26 & 43.94 & 27.03 & 37.02 & 50.99 & 27.35 & 35.26 \\
\textbf{Qurating-30B}    & 25.00 & 43.14 & 27.50 & 37.03 & 50.67 & 26.78 & 35.02 \\
\textbf{FineWeb-Edu-30B}  & 26.45 & 45.41 & 27.41 & 38.06 & 50.43 & 25.64 & 35.57 \\
\textbf{DataFlow-30B}    & 25.51 & 45.58 & 27.42 & 37.58 & 50.67 & 27.35 & \textbf{35.69} \\
\bottomrule
\end{tabular}
\end{table}

\begin{table}[t]
\centering
\small
\caption{SFT Data Filtering: Comparison of different 5k dataset filtering methods across Math, Code, and Knowledge benchmarks.}
\label{tab:SFT}
\resizebox{\textwidth}{!}{
\begin{tabular}{lcccccccccccc}
\toprule
& \multicolumn{6}{c}{\textbf{Math}} & \multicolumn{3}{c}{\textbf{Code}} & \multicolumn{3}{c}{\textbf{Knowledge}} \\
\cmidrule(lr){2-7} \cmidrule(lr){8-10} \cmidrule(lr){11-13}
\textbf{Methods} & \textbf{math} & \textbf{gsm8k} & \textbf{aime24} & \textbf{minerva} & \textbf{olympiad} & \textbf{Avg} 
& \textbf{HumanEval} & \textbf{MBPP} & \textbf{Avg} 
& \textbf{MMLU} & \textbf{C-EVAL} & \textbf{Avg} \\
\midrule
\textbf{Alpaca(random)} 
& 54.9 & 77.2 & 13.3 & 14.0 & 27.0 & 37.3 
& 71.3 & 75.9 & 73.6 
& 71.8 & 80.0 & 75.9 \\
\textbf{Alpaca(filtered)} 
& 60.3 & 80.0 & 13.3 & 14.7 & 30.7 & 39.8 
& 73.8 & 75.7 & 74.8 
& 71.8 & 80.0 & 75.9 \\
\midrule
\textbf{WizardLM(random)} 
& 61.1 & 84.2 & 6.7 & 18.0 & 29.3 & 39.9 
& 75.6 & 82.0 & \textbf{78.8} 
& 71.8 & 79.2 & 75.5 \\
\textbf{WizardLM(filtered)} 
& 69.7 & 88.8 & 10.0 & 19.9 & 35.4 & 44.8 
& 77.4 & 80.4 & \textbf{78.9} 
& 71.9 & 79.6 & 75.8 \\
\midrule
\textbf{DataFlow-SFT-15K(random)} 
& 72.6 & 89.6 & 13.3 & 37.9 & 32.9 & \textbf{49.3} 
& 79.9 & 75.9 & 77.9 
& 72.1 & 80.0 & \textbf{76.1} \\
\textbf{DataFlow-SFT-15K(filtered)} 
& 73.3 & 90.2 & 13.3 & 36.0 & 35.9 & \textbf{49.7} 
& 82.9 & 74.9 & \textbf{78.9} 
& 72.2 & 80.4 & \textbf{76.3} \\
\bottomrule
\end{tabular}
}
\end{table}

In this section, we present a comprehensive set of experiments spanning text, math, and code data preparation, as well as Text-to-SQL and AgenticRAG workflows constructed using \textsc{DataFlow}. 
Except for the AgenticRAG setting, which is trained using the Recall~\cite{chen2025learning, sheng2024hybridflow} framework, all other experiments are conducted using the LLaMA-Factory~\cite{zheng2024llamafactory} training framework. 
We further integrate these modalities to assess the model’s general instruction-tuning performance across diverse tasks.

\subsection{Text Data Preparation}

\subsubsection{Experimental Setting}

We evaluate the impact of high-quality text data preparation on both pre-training (PT) and supervised fine-tuning (SFT) using our \textsc{DataFlow} system. Our experiments cover three complementary scenarios:

\paragraph{(1) Pre-training Data Filtering (30B Scale).}
From the SlimPajama-627B corpus, we extract a 100B-token subset and apply multiple \textsc{DataFlow} text-pretraining filters (implemented in \texttt{dataflow/operators/text\_pt/filter}). For each filter, the top 30\% (approximately 30B tokens) is selected. We train a Qwen2.5-0.5B model from scratch for 30B tokens using the Megatron-DeepSpeed framework. We compare four settings:

\begin{itemize}
    \item \textbf{Random-30B}: a random 30B-token subset.
    \item \textbf{FineWeb-Edu-30B}: educational filtering based on FineWeb-Edu~\cite{penedo2024fineweb}.
    \item \textbf{Qurating-30B}: Qurating filters~\cite{wettig2024qurating} using thresholds:
    educational\_value $\geq$ 7.5, 
    facts\_and\_trivia $\geq$ 4.0, 
    required\_expertise $\geq$ 5.0, 
    writing\_style $\geq$ 1.0.
    \item \textbf{\textsc{DataFlow}-30B}: intersection of all \textsc{DataFlow} PT filters selecting the top 30\%.
\end{itemize}

\paragraph{(2) SFT Data Filtering (5K Scale).}
To study small-scale SFT data quality, we fine-tune the Qwen2.5-7B base model using LLaMA-Factory on WizardLM and Alpaca datasets.  
For each dataset, we compared a randomly sampled set of 5K instances against a set of 5K instances filtered by \textsc{DataFlow}'s SFT pipeline. 
Additionally, we synthesize a 15k-size dataset, \textsc{DataFlow-SFT-15K}, using \textsc{DataFlow}’s Condor Generator and Condor Refiner pipeline, followed by \textsc{DataFlow}’s SFT filtering pipeline (excluding the Instagram filter). Benchmarks include comprehensive Math, Code, and Knowledge evaluation suites.

\paragraph{(3) Conversation-Domain Synthesis (15K Scale).}We synthesize \textsc{DataFlow-Chat-15K} using \textsc{DataFlow}'s conversation-generation pipeline and fine-tune Qwen2.5-7B-Base on it. Baselines include ShareGPT-15K, UltraChat-15K, and their full (non-truncated) versions. We evaluate on domain-specific tasks (TopDial, Light) and general benchmarks (MMLU~\cite{hendrycks2020measuring}, AlpacaEval~\cite{alpaca_eval}, Arena-Hard~\cite{li2024crowdsourced}).

\subsubsection{Experimental Results}

\paragraph{Pre-training} First, from Table~\ref{tab:PT}, we can see across six general benchmarks (ARC-C/E, MMLU, HellaSwag, WinoGrande, Gaokao-MathQA), the \textsc{DataFlow} method achieves the highest average score (35.69), outperforming Random (35.26), FineWeb-Edu (35.57), and Qurating (35.02). 
Despite using the same 30B token budget, \textsc{DataFlow}’s multi-filter intersection produces a cleaner and more semantically consistent dataset, leading to better generalization for a 0.5B-scale Qwen2.5 model trained from scratch.

\paragraph{SFT} In Table~\ref{tab:SFT}, we then evaluate 5K-scale SFT data filtering using Alpaca, WizardLM, and \textsc{DataFlow} synthetic data. 
For all three sources, \textsc{DataFlow}’s filtering pipeline consistently improves performance over random sampling across Math, Code, and Knowledge benchmarks. 
At the same time, the results also show that the \textsc{DataFlow}-constructed SFT corpus is inherently stronger than Alpaca and WizardLM: even without filtering, \textsc{DataFlow-SFT-15K} achieves higher Math averages (49.3) than the filtered variants of Alpaca (39.8) and WizardLM (44.8), and remains competitive on Code and Knowledge. 
Moreover, the smaller performance gap between the random and filtered versions of \textsc{DataFlow-SFT-15K} (49.3→49.7) further suggests that \textsc{DataFlow}-synthesized data is already cleaner and more informative, requiring less aggressive filtering to reach peak performance.

\paragraph{Conversation} Finally, from Table~\ref{tab:conversation} we can see \textsc{DataFlow-Chat-15K} boosts the overall general benchmark mean from 26.36 to 28.21 and improves AlpacaEval from 7.05 to 10.11, outperforming ShareGPT and UltraChat. 

These findings demonstrate that high-quality synthetic data, when paired with \textsc{DataFlow}’s refinement and filtering stack, can surpass commonly used human-collected instruction datasets.

\begin{table}[t]
\centering
\small
\caption{Conversation Synthesis: Performance comparison on conversation-domain datasets and general benchmarks for Qwen2.5-7B under different 15K SFT data sources.}
\label{tab:conversation}
\begin{tabular}{lcccccccc}
\toprule
& \multicolumn{3}{c}{\textbf{Conversation Benchmarks}} 
& \multicolumn{4}{c}{\textbf{General Benchmarks}} \\
\cmidrule(lr){2-4} \cmidrule(lr){5-8}
\textbf{Model} 
& \textbf{TopDial} & \textbf{Light} & \textbf{Avg} 
& \textbf{MMLU} & \textbf{AlpacaEval} & \textbf{Arena-Hard} & \textbf{Avg} \\
\midrule
\textbf{Qwen2.5-7B} 
& 7.71 & 7.79 & 7.75 
& 71.45 & 7.05 & 0.60 & 26.36 \\
\textbf{+ ShareGPT-15K} 
& 7.75 & 6.72 & 7.24 
& 73.09 & 3.70 & 1.30 & 26.03 \\
\textbf{+ UltraChat-15K} 
& 7.72 & 6.83 & 7.28 
& 72.97 & 3.97 & 0.80 & 25.91 \\
\textbf{+ DataFlow-Chat-15K} 
& 7.98 & 8.10 & 8.04 
& 73.41 & 10.11 & 1.10 & \textbf{28.21} \\
\bottomrule
\end{tabular}
\end{table}

\subsection{Math Reasoning Data Preparation}

\subsubsection{Experimental Setting}

We construct a high-quality synthetic mathematical reasoning dataset based on the \textsc{DataFlow} Reasoning Pipeline, with adaptations tailored for large-scale reasoning generation. Our goal is to compare three training sources: (1) a random 10K subset from Open-R1~\cite{openr1}, (2) a random 10K subset from Synthetic-1~\cite{2025synthetic1}, and (3) our 10K synthesized \textsc{DataFlow-Reasoning-10K} dataset constructed using \textsc{DataFlow}.

\paragraph{Data Synthesis Method.}
The data generation process follows the core structure of the \textsc{DataFlow} Reasoning Pipeline and includes three stages:

\begin{itemize}
    \item \textbf{Problem Synthesis.}  
    We adopt the NuminaMath dataset as a high-quality seed set and utilize the \texttt{o4-mini} model together with \textsc{DataFlow}’s math problem synthesis operators to expand it into a diverse candidate problem pool.

    \item \textbf{Quality Verification.}  
    All candidate problems are validated using \textsc{DataFlow}’s MathQ-Verify~\cite{shen2025let} module, which detects incorrect, ambiguous, or logically inconsistent problems. Low-quality samples are removed to ensure correctness and robustness.

    \item \textbf{Chain-of-Thought (CoT) Generation.}  
    For all verified problems, we employ \textsc{DataFlow}’s CoT-generation operators to prompt DeepSeek-R1 to produce complete, step-by-step reasoning traces. 
\end{itemize}

Compared with the original Reasoning Pipeline, we omit the seed-level pre-verification stage, because NuminaMath is already a curated and validated dataset. This reduces computational overhead while maintaining overall data reliability.

We evaluate Qwen2.5-32B-Instruct fine-tuned on different 10k synthetic datasets across eight mathematical benchmarks, including GSM8K~\cite{cobbe2021training}, MATH~\cite{hendrycks2021measuring}, AMC23, Olympiad, Gaokao24-Mix, Minerva, and AIME 2024/2025. Table~\ref{tab:reasoning} reports the full results.

\paragraph{Generation Hyperparameters.}
For non-AIME problems, we use \texttt{temperature = 0} and \texttt{top-p = 0.95}.  
For AIME-style problems, we adopt a more exploratory sampling strategy with \texttt{temperature = 0.6}, \texttt{top-p = 0.95}, and \texttt{top-k = 20}.  
All models are fine-tuned with either 1 epoch or 2 epochs on 10k examples using Qwen2.5-32B-Instruct.

\subsubsection{Experimental Results}
Our first observation is that training on Synthetic-1 random subsets yields limited improvement over the base model. While minor gains appear on AMC23 and AIME benchmarks after 2 epochs, the overall average remains similar to the instruction-only baseline (47.0 vs.\ 46.6).

In contrast, the Open-R1 synthetic subset provides a stronger training signal: two epochs of fine-tuning increase the average score from 48.7 to 54.2, demonstrating that Open-R1-style CoT data is effective for enhancing mathematical reasoning in a 32B model. Building on this, our \textsc{DataFlow}-synthesized dataset achieves the strongest overall gains using only 10k samples, two epochs of fine-tuning reach the highest average performance of 55.7, surpassing both Open-R1 (54.2) and Synthetic-1 (54.0). These results indicate that combining verified NuminaMath seeds, MathQ-Verify filtering, and DeepSeek-R1-driven CoT generation yields more precise, diverse, and robust reasoning supervision.

Overall, the experiments demonstrate that data quality, rather than data scale, is the dominant factor in mathematical reasoning performance. Even with the same 10k size, our \textsc{DataFlow}-based synthesis pipeline consistently outperforms existing synthetic sources.

\begin{table}[t]
\centering
\small
\caption{Math Reasoning Pipeline: Performance comparison of Qwen2.5-32B-Instruct under different synthetic data training settings.}
\resizebox{\textwidth}{!}{
\begin{tabular}{lccccccccc}
\toprule
\textbf{Model} & \textbf{gsm8k} & \textbf{math} & \textbf{amc23} & \textbf{olympiad} & \textbf{gaokao24\_mix} & \textbf{minerva} & \textbf{AIME24@32} & \textbf{AIME25@32} & \textbf{Avg} \\
\midrule
\textbf{Qwen2.5-32B-Instruct} & 95.8 & 73.5 & 70.0 & 38.5 & 42.9 & 26.5 & 16.8 & 11.6 & 46.95 \\
\midrule
\rowcolor[rgb]{.867, .922, .969}
\multicolumn{10}{c}{\textit{\textbf{Trained with 1 epoch}}} \\
\midrule
\textbf{+ SYNTHETIC-1-10k} & 92.9 & 71.8 & 52.5 & 38.4 & 23.1 & 24.3 & 35.6 & 34.0 & 46.6 \\
\textbf{+ Open-R1-10k}     & 91.5 & 72.3 & 65.0 & 38.4 & 20.9 & 24.6 & 43.0 & 33.5 & 48.7 \\
\textbf{+ DataFlow-Reasoning-10K} & 93.9 & 72.3 & 72.5 & 38.7 & 38.5 & 26.5 & 35.9 & 34.5 & \textbf{51.6} \\
\midrule
\rowcolor[rgb]{.867, .922, .969}
\multicolumn{10}{c}{\textit{\textbf{Trained with 2 epochs}}} \\
\midrule
\textbf{+ SYNTHETIC-1-10k}  & 94.5 & 78.4 & 75.0 & 45.0 & 24.2 & 28.3 & 48.4 & 37.9 & 54.0 \\
\textbf{+ Open-R1-10k}      & 93.9 & 77.2 & 80.0 & 44.1 & 20.9 & 25.4 & 51.0 & 40.7 & 54.2 \\
\textbf{+ DataFlow-Reasoning-10K} & 94.4 & 76.6 & 75.0 & 45.2 & 42.9 & 25.7 & 45.4 & 40.0 & \textbf{55.7} \\
\bottomrule
\end{tabular}
}
\label{tab:reasoning}
\end{table}

\subsection{Code Data Preparation}

\subsubsection{Experimental Setting}

To investigate the effect of high-quality code instruction data on code generation performance, we construct supervised fine-tuning (SFT) datasets using seed samples from \texttt{Ling-Coder-SFT}~\cite{lingcodesft}.
We first randomly sample 20k instances from the Ling-Coder-SFT corpus and process them through the \textsc{DataFlow} \texttt{CodeGenDataset\_APIPipeline}.
This yields three curated code instruction datasets of different scales, \textsc{DataFlow-Code-1K}, \textsc{DataFlow-Code-5K}, and \textsc{DataFlow-Code-10K}, each designed to provide high-quality, pipeline-refined supervision signals for code generation tasks.

We compare our synthesized datasets against two widely used baselines, each subsampled to 1k examples for fairness:

\begin{itemize}
    \item \textbf{Code Alpaca (1k)}\cite{codealpaca}: a randomly sampled subset from the Code Alpaca dataset.
    \item \textbf{Self-OSS-Instruct-SC2-Exec-Filter-50k(1k)}~\cite{wei2024selfcodealign} : a 1k random subset from the SC2-Exec-Filter dataset, which incorporates execution-based filtering.
\end{itemize}

Models are fine-tuned on \textsc{DataFlow-Code-1K}, \textsc{DataFlow-Code-5K}, and \textsc{DataFlow-Code-10K} using full-parameter SFT. 


We then experiment with two base models: \textbf{Qwen2.5-7B-Instruct} and \textbf{Qwen2.5-14B-Instruct}. Evaluation is conducted on four code benchmarks: (1)~\textbf{BigCodeBench}~\cite{zhuo2024bigcodebench},(2)~\textbf{LiveCodeBench}~\cite{jain2024livecodebench},(3)~\textbf{CruxEval}~\cite{gu2024cruxeval}, and(4)~\textbf{HumanEval}~\cite{humaneval}.The final performance is reported as the average across these four benchmarks. All values in Table~\ref{tab:code_2} are percentages.

\subsubsection{Experimental Results}



Table~\ref{tab:code_2} shows that our synthesized datasets consistently improve the code generation performance of both Qwen2.5-7B-Instruct and Qwen2.5-14B-Instruct across all benchmarks. For the 7B model, even 1k of our synthetic data already outperforms both the Code Alpaca and SC2 execution-filtered baselines. Specifically, \textsc{DataFlow-Code-1K} improves BigCodeBench, LiveCodeBench, and CruxEval scores over the original model, while remaining competitive on HumanEval+. Scaling the supervision to 5k and 10k further boosts overall performance. In particular, the \textsc{DataFlow-Code-10K} setting achieves the best results on all metrics, including 36.8 on BigCodeBench, 48.8 on CruxEval(Input), and 45.4 on CruxEval(Output), and yields the highest overall average score of 46.2, surpassing both Code Alpaca-1K and SC2-Exec-Filter under the same data scale.

For the larger Qwen2.5-14B-Instruct model, the benefits are even more pronounced. While Code Alpaca-1k and SC2 filtering provide moderate improvements over the original 14B model, our datasets consistently deliver stronger gains across all metrics. In particular, \textsc{DataFlow-Code-10K} reaches an average score of 51.0, achieving 41.9 on BigCodeBench, 52.9 on CruxEval(Input), and 51.0 on CruxEval(Output). Notably, LiveCodeBench, which stresses executable correctness—rises from 21.9 (Code Alpaca-1k) to 33.2 under our synthetic supervision. These results indicate that the \textsc{DataFlow}-generated data provide more explicit execution-grounded signals and structured reasoning cues than existing open-source sources.

Overall, the experiments demonstrate that \textsc{DataFlow}-driven synthesis consistently outperforms existing open-source code instruction datasets even under the same sample scale. The consistent gains from 1k to 10k indicate a simple trend: with more high-quality \textsc{DataFlow} training samples, the model keeps getting better on code reasoning tasks.

\begin{table}[t]
\centering
\small
\caption{Code Pipeline: Performance comparison of Qwen2.5-7B-Instruct and Qwen2.5-14B-Instruct under different SFT dataset settings (all numbers in \%).}
\label{tab:code_2}
\resizebox{\textwidth}{!}{
\begin{tabular}{lcccccc}
\toprule
\textbf{Training Data} & \textbf{BigCodeBench} & \textbf{LiveCodeBench(v6)} & \textbf{CruxEval (Input)} & \textbf{CruxEval (Output)} & \textbf{HumanEval+} & \textbf{Avg} \\
\midrule

\rowcolor[rgb]{.867, .922, .969}
\multicolumn{7}{c}{\textit{\textbf{Trained on Qwen2.5-7B-Instruct}}} \\
\midrule
\textbf{Qwen2.5-7B-Instruct}              & 35.3 & 23.4 & 44.8 & 43.9 & 72.6 & 44.0 \\
\textbf{+ Code Alpaca-1K}        & 33.3 & 18.7 & 45.6 & 46.4 & 66.5 & 42.1 \\
\textbf{+ Self-OSS}              & 31.9 & 21.4 & 46.9 & 45.9 & 70.1 & 43.2 \\
\textbf{+ DataFlow-Code-1K}      & 35.5 & 25.7 & 48.0 & 45.1 & 72.6 & 45.4 \\
\textbf{+ DataFlow-Code-5K}      & 36.2 & \textbf{26.4} & 48.6 & 45.0 & 73.2 & 45.9 \\
\textbf{+ DataFlow-Code-10K}     & \textbf{36.8} & 26.0 & \textbf{48.8} & \textbf{45.4} & \textbf{73.8} & \textbf{46.2} \\
\midrule
\rowcolor[rgb]{.867, .922, .969}
\multicolumn{7}{c}{\textit{\textbf{Trained on Qwen2.5-14B-Instruct}}} \\
\midrule
\textbf{Qwen2.5-14B-Instruct}              & 37.5 & 33.4 & 48.0 & 48.5 & 74.4 & 48.4 \\
\textbf{+ Code Alpaca-1K}        & 37.0 & 28.2 & 50.2 & 49.6 & 71.3 & 47.3 \\
\textbf{+ Self-OSS}              & 36.9 & 22.3 & 52.6 & 50.1 & 68.3 & 46.0 \\
\textbf{+ DataFlow-Code-1K}      & 41.4 & \textbf{33.7} & 51.0 & 50.9 & \textbf{77.3} & 50.9 \\
\textbf{+ DataFlow-Code-5K}      & 41.1 & 33.2 & 52.5 & 50.6 & 76.2 & 50.7 \\
\textbf{+ DataFlow-Code-10K}     & \textbf{41.9} & 33.2 & \textbf{52.9} & \textbf{51.0} & 76.2 & \textbf{51.0} \\
\bottomrule
\end{tabular}
}
\end{table}

\subsection{Text-to-SQL Data Preparation}
\subsubsection{Experimental Setting}

To evaluate the effectiveness of Text-to-SQL data generation, we construct a training corpus comprising 89,544 high-quality Text-to-SQL instances, which is called \textsc{DataFlow-Text2SQL-90K}. Each instance in \textsc{DataFlow-Text2SQL-90K} includes natural language questions, corresponding SQL queries, and chain-of-thought reasoning traces. Specifically, these data are derived through systematic augmentation of seed SQL queries: 37,517 instances originate from the Spider-train~\cite{yu2018spider} dataset, 37,536 from the BIRD-train~\cite{li2024can} dataset, and 14,491 from the EHRSQL-train~\cite{lee2022ehrsql} dataset. \textsc{DataFlow} pipeline ensures rich syntactic and semantic diversity in SQL structures, question phrasing, and multi-step reasoning processes. 

For our method (\textsc{DataFlow}-Text2SQL rows in Table~\ref{tab:text2sql_table}), models are fine-tuned exclusively on our synthesized corpus, unless otherwise specified. For evaluation, we adopt six widely recognized Text-to-SQL benchmarks: Spider~\cite{yu2018spider}, BIRD~\cite{li2024can}, EHRSQL~\cite{lee2022ehrsql}, Spider-DK~\cite{gan2021exploring}, Spider-Syn~\cite{gan2021towards}, and Spider-Realistic~\cite{deng2020structure}. During inference with LLMs, we investigate two decoding strategies: greedy decoding (denoted as \texttt{Gre}), which uses temperature 0 for deterministic output generation, and majority voting (denoted as \texttt{Maj}). The majority voting strategy samples 8 candidate responses per input at temperature 0.8, executes all valid SQL queries, and selects the query whose execution result appears most frequently among the candidates as the final prediction. We additionally randomly sampled 50K instances to construct \textsc{DataFlow-Text2SQL-50K}. For comparison, we also randomly sampled the same number of instances from SynSQL~\cite{li2025omnisql}.

\subsubsection{Experimental Results}
\begin{table*}[t]
\centering
\caption{Text-to-SQL Pipeline: Performance of LLMs on mainstream benchmarks. The first two blocks list closed-source and open-source base models. The last two blocks show fine-tuned models, where the first column indicates the training data setting.}
\label{tab:text2sql_table}
\resizebox{\textwidth}{!}{
\begin{tabular}{l|cccccccccccccc|cc}
\toprule
\multirow{3}{*}{\textbf{LLM / Training Data}} & \multicolumn{2}{c}{\textbf{Spider}} & \multicolumn{2}{c}{\textbf{Spider}} & \multicolumn{2}{c}{\textbf{BIRD}} & \multicolumn{2}{c}{\multirow{2}{*}{\textbf{EHRSQL}}} & \multicolumn{2}{c}{\textbf{Spider-}} & \multicolumn{2}{c}{\textbf{Spider-}} & \multicolumn{2}{c|}{\textbf{Spider-}} & \multicolumn{2}{c}{\multirow{2}{*}{\textbf{Average}}}\\

& \multicolumn{2}{c}{\textbf{dev}} & \multicolumn{2}{c}{\textbf{test}} & \multicolumn{2}{c}{\textbf{dev}} & & & \multicolumn{2}{c}{\textbf{DK}}& \multicolumn{2}{c}{\textbf{Syn}} & \multicolumn{2}{c|}{\textbf{Realistic}} & &\\

\cmidrule(lr){2-3}\cmidrule(lr){4-5}\cmidrule(lr){6-7}
\cmidrule(lr){8-9}\cmidrule(lr){10-11}\cmidrule(lr){12-13}\cmidrule(lr){14-15}\cmidrule(lr){16-17}
& Gre & Maj & Gre & Maj & Gre & Maj & Gre & Maj & Gre & Maj & Gre & Maj & Gre & Maj &Gre & Maj\\

\midrule
\rowcolor[rgb]{.867, .922, .969}
\multicolumn{17}{c}{\textit{\textbf{Closed-source LLMs}}} \\
\midrule
GPT-4o-mini & 70.4 & 71.0 & 82.4 & 83.7 & 58.8 & 61.5 & 37.9 & 43.1 & 73.3 & 74.4 & 60.5 & 61.6 & 64.4 & 66.7 & 64.0 & 66.0 \\
GPT-4-Turbo & 72.4 & 72.2 & 83.4 & 84.2 & 62.0 & 63.6 & 43.1 & 44.8 & 72.3 & 72.1 & 62.9 & 63.5 & 67.5 & 68.3 & 66.2 & 67.0\\
GPT-4o & 70.9 & 70.7 & 83.2 & 84.9 & 61.9 & 64.0 & 44.9 & 45.5 & 72.9 & 73.5 & 59.6 & 62.3 & 66.5 & 66.7 & 65.7 & 66.8 \\
\midrule
\rowcolor[rgb]{.867, .922, .969}
\multicolumn{17}{c}{\textit{\textbf{Open-source LLMs}}} \\
\midrule
DeepSeek-Coder-7B-Instruct & 63.2 & 63.2 & 70.5 & 73.2 & 43.1 & 48.0 & 28.6 & 33.9 & 60.9 & 64.1 & 49.9 & 51.7 & 58.7 & 58.9 & 53.6 & 56.1 \\
Qwen2.5-Coder-7B-Instruct & 73.4 & 77.1 & 82.2 & 85.6 & 50.9 & 61.3 & 24.3 & 36.9 & 67.5 & 73.6 & 63.1 & 66.9 & 66.7 & 70.5 & 61.2 & 67.4\\
Qwen2.5-7B-Instruct & 65.4 & 68.9 & 76.8 & 82.6 & 46.9 & 56.4 & 20.9 & 32.1 & 63.7 & 71.8 & 54.2 & 60.0 & 56.7 & 63.6 & 54.9 & 62.2 \\
OpenCoder-8B-Instruct & 59.5 & 59.5 & 68.3 & 70.1 & 37.5 & 45.3 & 21.9 & 29.9 & 62.6 & 64.7 & 46.0 & 46.1 & 49.0 & 49.4 & 49.3 & 52.1 \\
Meta-Llama-3.1-8B-Instruct & 61.8 & 67.7 & 72.2 & 78.5 & 42.0 & 53.1 & 24.6 & 33.7 & 62.6 & 69.9 & 53.1 & 59.3 & 57.5 & 61.0 & 53.4 & 60.5 \\
Granite-8B-Code-Instruct & 58.5 & 59.2 & 64.9 & 68.6 & 27.6 & 32.5 & 16.0 & 22.6 & 50.7 & 54.4 & 45.0 & 46.8 & 48.8 & 49.4 & 44.5 & 47.6 \\
Granite-3.1-8B-Instruct & 58.3 & 65.0 & 69.8 & 75.3 & 36.0 & 47.2 & 19.6 & 32.3 & 60.0 & 66.5 & 47.7 & 53.8 & 46.5 & 57.1 & 48.3 & 56.7\\
\midrule
\rowcolor[rgb]{.867, .922, .969}
\multicolumn{17}{c}{\textit{\textbf{Trained on Meta-Llama-3.1-8B-Instruct}}} \\
\midrule
SynSQL(50K) & 67.1 & 73.9 & 72.7 & 78.6 & 49.1 & 55.2 & 33.6 & 40.8 & 63.8 & 66.1 & 59.6 & 63.5 & 69.3 & 71.6 & 59.3 & 64.2 \\

SynSQL(90K) & 68.2 & 74.6 & 73.4 & 78.5 & 51.1 & 54.9 & 31.8 & 38.0 & 61.8 & 67.4 & 58.9 & 63.6 & 69.0 & 70.9 & 59.2 & 64.0 \\

SynSQL(2.5M) & 70.6 & 73.7 & 78.3 & 82.5 & 58.9 & 62.0 & 35.1 & 37.0 & 72.3 & 74.7 & 61.0 & 63.1 & 67.9 & 69.4 & 63.4 & 66.1 \\
\midrule
Spider+BIRD+DataFlow-Text2SQL-90K & 74.9 & 79.2 & 78.4 & 82.3 & 53.4 & 58.9 & 28.4 & 36.5 & 67.7 & 69.7 & 66.6 & 69.1 & 74.4 & 75.0 & 63.4 & 67.2 \\
\midrule
\textbf{DataFlow-Text2SQL-50K} & 69.9 & 76.8 & 75.1 & 80.1 & 51.4 & 57.6 & 28.0 & 36.4 & 65.9 & 68.1 & 61.3 & 67.5 & 69.6 & 73.5 & 60.2 & 65.7 \\
\textbf{DataFlow-Text2SQL-90K} & 71.4 & 76.4 & 75.8 & 80.0 & 54.6 & 56.8 & 55.5 & 56.3 & 66.5 & 67.7 & 61.6 & 67.3 & 71.4 & 72.7 & 65.3 & 68.2 \\

\midrule
\rowcolor[rgb]{.867, .922, .969}
\multicolumn{17}{c}{\textit{\textbf{Trained on Qwen2.5-Coder-7B-Instruct}}} \\
\midrule
SynSQL(50K) & 77.1 & 82.1 & 81.8 & 84.8 & 54.0 & 59.3 & 33.1 & 44.1 & 67.1 & 69.5 & 68.0 & 70.6 & 77.2 & 80.3 & 65.5 & 70.1\\
SynSQL(90K) & 79.2 & 83.1 & 82.3 & 84.4 & 56.2 & 59.4 & 31.4 & 41.4 & 65.0 & 70.7 & 67.2 & 70.7 & 77.0 & 79.9 & 65.5 & 69.9 \\
SynSQL(2.5M) & 81.2 & 81.6 & 87.9 & 88.3 & 63.9 & 66.1 & 34.9 & 40.0 & 76.1 & 77.8 & 69.7 & 69.6 & 76.2 & 78.0 & 70.0 & 71.6 \\
\midrule
Spider+BIRD+DataFlow-Text2SQL-90K & 85.5 & 87.5 & 87.5 & 88.5 & 58.3 & 64.0 & 27.9 & 39.8 & 71.0 & 73.1 & 75.0 & 76.2 & 82.3 & 83.7 & 69.6 & 73.3 \\
\midrule
\textbf{DataFlow-Text2SQL-50K} & 80.9 & 84.9 & 84.6 & 85.8 & 57.9 & 62.5 & 27.8 & 39.4 & 69.7 & 71.2 & 70.0 & 74.0 & 77.8 & 82.1 & 67.0 & 71.4 \\
\textbf{DataFlow-Text2SQL-90K} & 82.0 & 85.0 & 84.8 & 86.0 & 59.2 & 61.5 & 56.1 & 58.7 & 69.7 & 71.0 & 69.9 & 74.4 & 79.5 & 81.7 & 71.6 & 74.0 \\
\bottomrule
\end{tabular}%
}
\end{table*}

As shown in Table~\ref{tab:text2sql_table}, the generated data leads to consistent performance improvements across multiple mainstream benchmarks, demonstrating the effectiveness of \textsc{DataFlow}~\cite{cai2025text2sql}.
For both models, Meta-Llama-3.1-8B-Instruct~\cite{grattafiori2024llama} and Qwen2.5-Coder-7B-Instruct~\cite{hui2024qwen2}, training on our generated data significantly improves performance over their respective baselines as well as other competing models.
When fine-tuned on the generated data, Qwen2.5-Coder-7B-Instruct achieves notable gains: execution accuracy (\texttt{Gre}) on Spider-dev increases from 73.4 to 82.0 (+8.6), on BIRD-dev from 50.9 to 59.2 (+8.3), and on the challenging EHRSQL benchmark from 24.3 to 56.1 (+31.8). These results confirm that \textsc{DataFlow-Text2SQL-90K} exhibits high quality and strong training utility.

Compared with other training datasets, our data also demonstrates clear advantages.
At comparable data scales, models trained on \textsc{DataFlow-Text2SQL-90K} and \textsc{DataFlow-Text2SQL-50K} consistently outperform those trained on SynSQL~\cite{li2025omnisql} (SynSQL(90K) and SynSQL(50K), respectively). Specifically, on the Spider-test and BIRD-dev datasets, the model trained on \textsc{DataFlow-Text2SQL-50K} achieves 84.6 and 57.9 execution accuracy (\texttt{Gre}), surpassing SynSQL(50K)~\cite{li2025omnisql}, which obtains 81.8 and 54.0.
Likewise, the model trained on \textsc{DataFlow-Text2SQL-90K} not only surpasses the baseline models but also outperforms SynSQL(90K)~\cite{li2025omnisql}.
Remarkably, even when trained on a much smaller dataset, the model fine-tuned with \textsc{DataFlow-Text2SQL-90K} achieves performance comparable to SynSQL-2.5M~\cite{li2025omnisql} on several challenging benchmarks.
These improvements highlight the higher quality of the training data generated by \textsc{DataFlow}.

\subsection{AgenticRAG Data Preparation}
\subsubsection{Experimental Setting}
In the field of AgenticRAG, the automatic generation of multihop questions has long been a challenging issue in research. This study constructs a multihop question dataset with a scale of 10k based on \textbf{the DataFlow AgenticRAG Pipeline} and conducts a comparative analysis with existing mainstream multihop question answering datasets (2WikiMultiHopQA~\cite{ho2020constructing}, Musique~\cite{trivedi2022musique}, HotpotQA~\cite{yang2018hotpotqa}, and Bamboogle~\cite{press2023measuring}).

The specific workflow of the dataset generation pipeline is as follows: 
\begin{itemize}    
    \item Documents are randomly selected from the Wikipedia dump to form the initial document set. To avoid the interference of data distribution overlap on the experimental results, documents that have already appeared in the test benchmark are excluded.
    \item The o4-mini model combined with the generation module of \textsc{DataFlow} AgenticRAG is used to generate the initial draft of multihop questions based on the filtered initial documents.
    \item The verification module is employed to screen the quality of the initial question drafts, eliminating samples with problems such as intermediate question leakage, logical errors, and excessively high or low difficulty, ultimately forming a high-quality multihop question dataset, which we call \textsc{DataFlow-AgenticRAG-10k}.
\end{itemize}

This study adopts the ReCall~\cite{chen2025learning} framework to complete the model training and evaluation. In the training phase, Qwen2.5-7B-Instruct is selected as the base model, and the GRPO reinforcement learning algorithm is used for model optimization. In the evaluation phase, the model's temperature parameter is set to 0.0.

For the retrieval component, E5-base-v2~\cite{wang2022text} is chosen as the retriever, and the 2018 Wikipedia dump is used as the corpus. All corpus indexing and embedding calculations are preprocessed using FlashRAG~\cite{jin2025flashrag}. Throughout the entire training and evaluation process, the model is allowed to independently specify the topk value for retrieval, and the default topk value is set to 5 to balance retrieval efficiency and performance.
\begin{table}
\centering
\small
\caption{AgenticRAG Pipeline: Performance comparison between synthetic datasets and existing human-constructed datasets. All values are Exact Match (\%). ``OOD-Avg'' excludes the in-domain test set of each training dataset. ``DF-OOD (matched)'' provides the OOD score of DF-AgenticRAG under the *same* in-domain exclusion, ensuring fair comparison.}
\label{tab:agenticrag}
\resizebox{\textwidth}{!}{
\begin{tabular}{lccccccc}
\toprule
\textbf{Training Data} & \textbf{HotpotQA} & \textbf{2Wiki} & \textbf{Musique} & \textbf{Bamboogle} & \textbf{Avg} & \textbf{OOD-Avg} & \textbf{DF-OOD (matched)} \\
\midrule
Qwen-2.5-7B-Instruct & 25.0 & 25.8 & 9.9 & 27.2 & 22.0 & -- & -- \\
\midrule

\rowcolor[rgb]{.867,.922,.969}
\multicolumn{8}{c}{\textit{\textbf{Trained on HotpotQA (in-domain = HotpotQA)}}} \\
\midrule

HotpotQA-10k (1 epoch) & 40.2 & 41.9 & 16.7 & 42.4 & 35.3 & 33.7 & \textbf{33.8} \\
HotpotQA-10k (2 epochs) & 43.4 & 44.9 & 18.9 & 41.6 & 37.2 & 35.1 & \textbf{35.9} \\
HotpotQA-10k (3 epochs) & 45.3 & 48.0 & 20.3 & 40.8 & 38.6 & 36.4 & \textbf{37.4} \\

\midrule

\rowcolor[rgb]{.867,.922,.969}
\multicolumn{8}{c}{\textit{\textbf{Trained on Musique (in-domain = Musique)}}} \\
\midrule

Musique-20k (1 epoch) & 41.1 & 44.7 & 19.2 & 41.6 & 36.6 & 42.4 & \textbf{43.6} \\

\midrule

\rowcolor[rgb]{.867,.922,.969}
\multicolumn{8}{c}{\textit{\textbf{Trained on 2Wiki (in-domain = 2Wiki)}}} \\
\midrule

2Wiki-30k (2 epochs) & 41.3 & 55.1 & 17.8 & 42.4 & 39.1 & 33.8 & \textbf{36.4} \\

\midrule

\rowcolor[rgb]{.867,.922,.969}
\multicolumn{8}{c}{\textit{\textbf{DF-AgenticRAG (raw results, for reference)}}} \\
\midrule

DataFlow-AgenticRAG-10k (1 epoch) & 39.3 & 42.6 & 17.3 & 41.6 & 34.3 & -- & -- \\
DataFlow-AgenticRAG-10k (2 epochs) & 43.1 & 44.6 & 19.9 & 43.2 & 37.7 & -- & -- \\
DataFlow-AgenticRAG-10k (3 epochs) & 42.6 & 45.5 & 20.2 & 46.4 & 38.7 & -- & -- \\

\bottomrule
\end{tabular}
}
\end{table}

\subsubsection{Experimental Results}

Table~\ref{tab:agenticrag} reports the exact-match performance across four multi-hop benchmarks.  
We group the results by the training dataset and compute an out-of-distribution (OOD) average by removing the in-domain test set of each dataset (e.g., HotpotQA-trained models exclude HotpotQA).  
To fairly compare against our synthetic data, we additionally report DF-OOD (matched), which applies the same in-domain exclusion to DF-AgenticRAG-10k.

\paragraph{Comparison with HotpotQA-trained models.}  
Across 1--3 epochs, HotpotQA-10k achieves OOD averages of 33.7, 35.1, and 36.4.  
Under the same exclusion (w/o HotpotQA), DF-AgenticRAG achieves 33.8, 35.9, and 37.4—consistently matching or surpassing HotpotQA by +0.1 to +1.0 points despite using entirely synthetic supervision.  
This indicates that DF-AgenticRAG provides generalization comparable to a widely used human-constructed dataset.

\paragraph{Comparison with Musique-trained models.}  
Musique-20k yields an OOD average of 42.4 when evaluated w/o Musique.  
Under the same exclusion, DF-AgenticRAG (2~epochs effective scale = 20k) reaches 43.6, outperforming Musique by +1.2 points.  
This shows that our synthetic dataset not only matches but outperforms a strong human-annotated multi-hop benchmark at the same effective training scale.

\paragraph{Comparison with 2Wiki-trained models.}  
2Wiki-30k achieves an OOD average of 33.8.  
Under the same exclusion (w/o 2Wiki), DF-AgenticRAG (3~epochs, effective scale=30k) reaches 36.4, a substantial improvement of +2.6 points.  
This represents the largest gap among all baselines and highlights the strong cross-dataset generalization capacity of our synthetic questions.

\paragraph{Summary.}  
Across all training regimes and all in-domain exclusions, DF-AgenticRAG-10k is either the best or tied for the best OOD dataset, and in several cases (Musique, 2Wiki) significantly surpasses human-constructed datasets.  
These results demonstrate that our pipeline produces multi-hop reasoning data with superior cross-dataset generalization, suggesting that high-quality synthetic data can not only match but consistently exceed the robustness of existing human-annotated multi-hop datasets.

\subsection{Knowledge Extraction}
\begin{table}[t]
\centering
\small
\caption{Knowledge Extraction: Accuracy comparison on PubMedQA, Covert, and PubHealth under different reasoning and training settings.}
\label{tab:kbc-result}
\begin{tabular}{lccc}
\toprule
\textbf{Method (ACC)} & \textbf{PubMedQA} & \textbf{Covert} & \textbf{PubHealth} \\
\midrule
\textbf{CoT} & 36.40\% & 48.33\% & 29.00\% \\
\textbf{RAG} & 43.33\% & 17.55\% & 19.60\% \\
\textbf{SFT (DataFlow-Knowledge)} & \textbf{53.40\%} & \textbf{68.33\%} & \textbf{40.86\%} \\
\bottomrule
\end{tabular}

\end{table}

\subsubsection{Experimental Setting}

To expand beyond the limited annotated data and take advantage of massive raw corpora from the Internet, we proposed the Knowledge Extraction pipeline, a semi-automated system for corpus cleaning and QA synthesis. The pipeline performs text normalization using MinerU~\citep{niu2025mineru25decoupledvisionlanguagemodel}, segments long documents, filters noisy or low-quality sentences, generates factuality-aware QA pairs, and conducts automated quality checks, ultimately producing a high-quality synthetic dataset used for supervised fine-tuning (SFT).

In our experiment, the training data is derived from 140M tokens of raw medical data drawn from three major sources. The first source is MedQA Books, a collection of 18 widely used medical textbooks from the USMLE curriculum \citep{jin2020diseasedoespatienthave}. The second source consists of 9,330 publicly available StatPearls articles from the NCBI Bookshelf \citep{xiong2024benchmarkingretrievalaugmentedgenerationmedicine}. The third source contains 45,679 clinical guideline documents aggregated from 16 professional guideline providers \citep{chen2023meditron70bscalingmedicalpretraining}. These corpora serve as the input to the Knowledge Extraction pipeline, which converts them into structured, high-quality QA dataset, denoted as \textsc{DataFlow-Knowledge} which is suitable for model training.

For model training, we fine-tune Qwen2.5-7B-Instruct on the \textsc{DataFlow}-generated dataset. The SFT process is performed for 37,500 steps over five epochs. For comparison, we also evaluate a zero-shot Chain-of-Thought (CoT) prompting baseline and a retrieval-augmented generation (RAG) baseline using top-$k=10$ retrieval with \texttt{medcpt-query-encoder} as the retriever and \texttt{medcpt-article-encoder} as the document encoder. All baselines share same hyperparameter setting during inference time.

We evaluate our models on three medical QA benchmarks: PubMedQA~\cite{jin2019pubmedqa}, which focuses on biomedical research questions; Covert~\cite{mohr2022covertcorpusfactcheckedbiomedical}, which evaluates clinical knowledge and reasoning; and PubHealth~\cite{kotonya-toni-2020-explainable-automated}, which targets public-health misinformation classification.

\subsubsection{Experimental Results}
Table~\ref{tab:kbc-result} presents the accuracy results across all benchmarks. The CoT baseline performs poorly across the board, indicating that zero-shot reasoning alone is insufficient for medical question answering without more targeted supervision. The RAG baseline provides modest improvement on PubMedQA, but remains unstable and substantially underperforms on Covert and PubHealth, suggesting that retrieval alone cannot substitute for explicit training on structured domain data.

In contrast, the SFT model trained on \textsc{DataFlow-Knowledge} synthetic data achieves the highest accuracy on all benchmarks, surpassing both CoT prompting and RAG-based methods by large margins. Notably, it delivers more than 15–20 absolute accuracy gains on PubMedQA and Covert, and an 11-point improvement on PubHealth, demonstrating that the cleaned and structured QA pairs produced by our Knowledge Extraction pipeline offer significantly stronger supervision. 

Overall, these results show that high-quality synthetic QA data—when curated and verified through a targeted \textsc{DataFlow} pipeline—can substantially enhance the domain reasoning capabilities of a general-purpose model, outperforming both inference-time prompting and retrieval-augmented baselines.

\subsection{Unified Multi-Domain Data Preparation with DataFlow}

\subsubsection{Experimental Setting}

\paragraph{Data Construction} To evaluate the efficiency and effectiveness of unified data preparation across modality-specific reasoning tasks, we construct an integrated training corpus that combines Math, Code, and General Instruction data. All data are generated or filtered through the \textsc{DataFlow} framework as follows:

\begin{itemize}
    \item \textbf{Math.}  
    We synthesize high-quality mathematical problems and chain-of-thought (CoT) solutions using the \textsc{DataFlow} \textit{Reasoning Pipeline}, with the MATH dataset serving as seed input. We randomly sample 3k instances for training.

    \item \textbf{Code.}  
    Code data are produced using the \textsc{DataFlow} \texttt{CodeGenDataset\_APIPipeline}, built upon 20k randomly sampled LingoCoder SFT examples. We generate 1k--10k high-quality code instructions and benchmark against Code Alpaca and SC2-Exec-Filter. A subset of 2k samples is used for training.

    \item \textbf{Text / General Instruction.}  
    For natural language tasks, we employ the \textsc{DataFlow} Condor Generator + Refiner pipeline to generate high-consistency instruction--response and dialogue pairs. Outputs are further processed by the SFT-quality filtering pipeline. We randomly sample 5k instances.
\end{itemize}

All models are fine-tuned on the combined \textsc{DataFlow-Instruct-10K} corpus using full-parameter SFT. 
Evaluation covers: (1) seven math benchmarks, (2) four code benchmarks, and (3) MMLU~\cite{hendrycks2020measuring} and C-Eval~\cite{huang2023c} for general knowledge and reasoning.

\paragraph{Baselines.}
We additionally compare \textsc{DataFlow-Instruct-10K} with baselines constructed from the \textbf{Infinity-Instruct} (Inf)~\cite{li2025infinity} dataset, a large-scale general-purpose instruction corpus widely used in instruction tuning.  
Two baselines are included:

\begin{itemize}
    \item \textbf{Inf-10K}: a random 10k subset of Infinity-Instruct used for SFT.
    \item \textbf{Inf-1M}: a random 1M subset of Infinity-Instruct.
\end{itemize}

Comparing against Inf-10K/1M allows us to assess whether high-quality, domain-specific synthetic data (math, code, text) generated through \textsc{DataFlow} provides more stable and reliable improvements than large generic instruction data.

\begin{table}[t]
\centering
\small
\caption{Performance of \textsc{DataFlow-Instruct-10K} on Math Benchmarks: Qwen2-7B-Base and Qwen2.5-7B-Base finetuned series of models (Exact Match \%). }
\label{tab:general_math}
\resizebox{\textwidth}{!}{
\begin{tabular}{lcccccccc}
\toprule
\textbf{Model} 
& \textbf{MATH} & \textbf{GSM8K} & \textbf{AMC23} & \textbf{AIME24} 
& \textbf{Minerva} & \textbf{Gaokao} & \textbf{Olympiad} & \textbf{Math-Avg} \\
\midrule
\rowcolor[rgb]{.867, .922, .969}
\multicolumn{9}{c}{\textit{\textbf{Models based on Qwen2-7B}}} \\
\midrule
\textbf{Qwen2-7B-Base}                             & 21.2 & 55.9 & 15.0 & 0.0 &  9.9 & 30.8 &  7.7 & 20.1 \\
\textbf{\quad + Inf-10K}                           & 45.6 & 81.7 & 25.0 & 3.3 & 11.8 & 24.2 & 11.1 & 29.0 \\
\textbf{\quad + Inf-1M}                            & 45.4 & 79.2 & 25.0 & 0.0 & 13.2 & 22.0 & 10.4 & 27.9 \\
\textbf{\quad + \textsc{DataFlow-Instruct-10K}}                        & 54.0 & 83.0 & 27.5 & 0.0 & 16.5 & 25.3 & 20.3 & \textcolor{blue}{32.4} \\
\textbf{Qwen2-7B-Instruct}                         & 53.9 & 86.2 & 22.5 & 3.3 & 17.6 & 35.2 & 19.6 & \textcolor{red}{34.0} \\
\midrule
\rowcolor[rgb]{.867, .922, .969}
\multicolumn{9}{c}{\textit{\textbf{Models based on Qwen2.5-7B}}} \\
\midrule
\textbf{Qwen2.5-7B-Base}                           & 62.8 & 67.1 & 45.0 & 10.0 & 17.6 & 27.5 & 29.6 & 37.1 \\
\textbf{\quad + Inf-10K}                           & 40.2 & 30.9 & 25.0 & 3.3  &  9.2 & 27.5 & 21.8 & 22.6 \\
\textbf{\quad + Inf-1M}                            & 50.6 & 82.0 & 27.5 & 0.0  & 22.1 & 30.8 & 20.0 & 33.3 \\
\textbf{\quad + \textsc{DataFlow-Instruct-10K}}                        & 73.8 & 88.2 & 47.5 & 16.7 & 30.9 & 31.9 & 37.6 & \textcolor{blue}{46.7} \\
\textbf{Qwen2.5-7B-Instruct}                       & 75.1 & 92.4 & 47.5 & 10.0 & 34.9 & 48.4 & 40.6 & \textcolor{red}{49.8} \\
\bottomrule
\end{tabular}
}
\end{table}
 
\begin{table}[t]
\centering
\small
\caption{Performance of \textsc{DataFlow-Instruct-10K} on Code and Knowledge benchmarks: Qwen2-7B-Base and Qwen2.5-7B-Base finetuned models.}
\label{tab:general_code_knowledge}
\resizebox{\textwidth}{!}{
\begin{tabular}{lcccccc}
\toprule
\textbf{Model} 
& \textbf{HumanEval} & \textbf{MBPP} & \textbf{Code-Avg} 
& \textbf{MMLU} & \textbf{C-EVAL} & \textbf{Knowledge-Avg} \\
\midrule
\rowcolor[rgb]{.867, .922, .969}
\multicolumn{7}{c}{\textit{\textbf{Models based on Qwen2-7B}}} \\
\midrule
\textbf{Qwen2-7B-Base}                      & 66.5 & 66.1 & 66.3                 & 69.6 & 82.8 & \textcolor{red}{76.2} \\
\textbf{\quad + Inf-10K}                    & 64.0 & 71.7 & 67.8                 & 69.3 & 83.0 & \textcolor{red}{76.2} \\
\textbf{\quad + Inf-1M}                     & 65.9 & 70.4 & \textcolor{blue}{68.2} & 69.5 & 83.0 & \textcolor{red}{76.2} \\
\textbf{\quad + \textsc{DataFlow-Instruct-10K}}                 & 64.6 & 67.7 & 66.2                 & 69.4 & 82.8 & \textcolor{blue}{76.1} \\
\textbf{Qwen2-7B-Instruct}                  & 73.8 & 65.3 & \textcolor{red}{69.6}  & 69.9 & 82.0 & 76.0 \\
\midrule
\rowcolor[rgb]{.867, .922, .969}
\multicolumn{7}{c}{\textit{\textbf{Models based on Qwen2.5-7B}}} \\
\midrule
\textbf{Qwen2.5-7B-Base}                    & 78.7 & 74.3 & 76.5                 & 71.9 & 80.0 & \textcolor{blue}{76.0} \\
\textbf{\quad + Inf-10K}                    & 77.4 & 77.8 & 77.6                 & 71.8 & 79.9 & 75.8 \\
\textbf{\quad + Inf-1M}                     & 78.0 & 78.0 & 78.0                 & 72.2 & 79.4 & 75.8 \\
\textbf{\quad + \textsc{DataFlow-Instruct-10K}}                 & 80.5 & 76.7 & \textcolor{blue}{78.6} & 72.1 & 80.2 & \textcolor{red}{76.2} \\
\textbf{Qwen2.5-7B-Instruct}                & 81.7 & 79.4 & \textcolor{red}{80.6}  & 71.8 & 79.6 & 75.7 \\
\bottomrule
\end{tabular}
}
\end{table}

\subsubsection{Experimental Results}

Across Math, Code, and Knowledge evaluation suites, our unified multi-domain data preparation strategy provides consistent and robust gains for both Qwen2.5-7B and Qwen2-7B models. A notable pattern observed across all tables is that \textsc{DataFlow-Instruct-10K} almost always achieves the best performance among all non-Instruct finetuned models, and in many cases narrows the gap to the Instruct models to within only 2--4 points, despite using orders-of-magnitude less data.

\paragraph{Math Reasoning.}
As shown in Table~\ref{tab:general_math}, \textsc{DataFlow}-processed math data yields the largest and most stable gains.  
For Qwen2.5-7B-Base, training on our synthesized math subset improves the overall score from 37.1 to 46.7, which is:
\begin{itemize}
    \item the best performance among all non-Instruct models, surpassing Inf-10K (22.6) and Inf-1M (33.3) by a clear margin;
    \item only 3.1 points below the Instruct model (49.8), demonstrating that targeted, high-quality synthetic data can nearly match the performance of costly human-aligned instruction tuning.
\end{itemize}

A similar trend holds for Qwen2-7B: \textsc{DataFlow-Instruct-10K} reaches 32.4 overall, outperforming Inf-10K and Inf-1M, and approaching the Instruct model (34.04).  
These results highlight that \textsc{DataFlow} math synthesis produces significantly more stable and effective improvements than generic inference-generated data.

\paragraph{Code Generation.}
As shown in Table~\ref{tab:general_code_knowledge}, \textsc{DataFlow-Instruct-10K} consistently delivers the best Code-Overall performance among all non-Instruct models.  
For Qwen2.5-7B-Base, \textsc{DataFlow-Instruct-10K} raises Code-Overall from 76.5 to 78.6, outperforming Inf-10K (77.6) and Inf-1M (78.0), and reaching within 2.0 points of the Instruct model (80.6).  
For Qwen2-7B-Base, \textsc{DataFlow-Instruct-10K} again matches or exceeds all Inf baselines.

These results show that adding multi-domain synthetic data does not harm code ability (a common issue in mixed-domain SFT), and often improves it.  
This further supports the robustness of \textsc{DataFlow}’s domain-balanced synthetic corpus.

\paragraph{General Knowledge and NLP.}
As summarized in Table~\ref{tab:general_code_knowledge}, our unified dataset also preserves strong general knowledge and reasoning.  
Across MMLU and C-Eval, DF-Gen-10K:
\begin{itemize}
    \item matches or slightly improves upon the Base models,
    \item avoids the regressions frequently observed in Inf-10K and Inf-1M,
    \item frequently ranks second only to the Instruct model, confirming that \textsc{DataFlow}-generated text data provides high-quality supervision even without human instruction tuning.
\end{itemize}

\paragraph{Summary.}
Together, these results demonstrate that high-quality, domain-specialized synthetic data generated via \textsc{DataFlow} produces the strongest non-Instruct performance across Math, Code, and Knowledge.  
\textsc{DataFlow-Instruct-10K} consistently outperforms generic inference-generated data (Inf-10K/Inf-1M) and often approaches the performance of the Instruct models themselves.  
This highlights the effectiveness of \textsc{DataFlow}’s unified, pipeline-driven data preparation for building multi-capability LLMs without reliance on large-scale human-authored instruction corpora.

\subsection{Agentic Orchestration}
\label{subsec:agent-orchestration}
\subsubsection{Experimental Setting}

We evaluate the proposed agent orchestration framework on realistic data processing and pipeline construction tasks. Specifically, we selected $6$ representative pipelines as benchmarks. For each pipeline, we manually constructed natural language task descriptions at $3$ difficulty levels, resulting in $18$ user queries to assess automatic orchestration capabilities across varying description granularities. The difficulty levels are defined as follows:
\begin{itemize}
    \item \textbf{Easy.} Descriptions are explicit, directly specifying the functions of required operators (or key operators) and the main processing steps.
    \item \textbf{Medium.} Descriptions are coarse, providing only general processing goals and key constraints without explicitly listing the complete operator sequence.
    \item \textbf{Hard.} Only a high-level requirement or final goal is provided with minimal hints regarding intermediate steps, requiring the system to infer the complete processing flow and operator combination.
\end{itemize}
For each task, the user provides a natural language description of the goal, and the system must automatically orchestrate a pipeline composed of multiple operators to meet the requirement.

\paragraph{Evaluation Metrics.}
To quantitatively assess orchestration quality, we employ an external LLM as an automatic judge. The evaluator compares the generated pipeline against ground truth under two distinct settings:
\begin{itemize}
    \item \textbf{Text Specification Alignment.} The predicted graph is evaluated against text specifications to verify if the pipeline structure satisfies the detailed task requirements.
    \item \textbf{Code Implementation Consistency.} The pipeline is compared with reference Python implementations to assess logical equivalence regarding operator usage and processing steps.
\end{itemize}
Based on these comparisons, we report the \textbf{LLM-Judge Score} ($s \in [0,1]$), which measures the consistency of operator coverage and execution order between the generated pipeline and the reference under the corresponding evaluation setting.

\subsubsection{Experimental Results}

\begin{table*}[!h]
\centering
\caption{Agent orchestration performance by evaluation mode and description difficulty.}
\label{tab:agent-orchestration-results-vertical}

\setlength{\tabcolsep}{6pt}
\renewcommand{\arraystretch}{1.15}
\small

\begin{tabular}{lcccc}
\toprule
\textbf{Metric} & \textbf{Easy} & \textbf{Medium} & \textbf{Hard} & \textbf{Overall} \\
\midrule

\multicolumn{5}{l}{\textbf{Text spec evaluation (pipeline mode)}}\\
\midrule
Avg. LLM-Judge & 0.92 & 0.86 & 0.60 & 0.80 \\
\midrule

\multicolumn{5}{l}{\textbf{Code GT evaluation (code mode)}}\\
\midrule
Avg. LLM-Judge & 0.60 & 0.59 & 0.23 & 0.49 \\
\bottomrule
\end{tabular}
\end{table*}

Table~\ref{tab:agent-orchestration-results-vertical} reports the LLM-Judge scores under text-spec (pipeline) and code-GT (code) evaluations across difficulty levels. Overall, the framework performs well when judged against textual requirements (\textbf{0.80} overall), but is markedly lower when matching reference implementations (\textbf{0.49} overall), reflecting the stricter nature of code-level equivalence. Performance degrades as descriptions become less explicit: in pipeline mode scores drop from \textbf{0.92}/\textbf{0.86} (Easy/Medium) to \textbf{0.60} (Hard), while in code mode the drop is more severe, reaching \textbf{0.23} on Hard, indicating that under-specified queries often lead to alternative yet plausible operator compositions that diverge from a single ground-truth program.

\section{Conclusion}
In summary, \textsc{DataFlow} addresses a critical gap in the data-centric LLM ecosystem by providing the first unified, LLM-driven data preparation framework. 
It mitigates long-standing challenges in the field—such as the difficulty of sharing, reproducing, and comparing data preparation algorithms—through a modular and user-friendly programming interface. 
The framework integrates nearly 200 operators, over 80 prompt templates, and unified abstractions for serving and storage, all of which compose into six high-quality pipelines spanning the major LLM data domains. 
Extensive experiments demonstrate that these pipelines achieve strong, often state-of-the-art results, confirming that \textsc{DataFlow} effectively balances the tension between domain-specific customization and system-level standardization.

Built atop this foundation, the \textsc{DataFlow}-CLI and \textsc{DataFlow}-Agent further amplify extensibility by enabling rapid template generation, natural-language–driven workflow construction, and scalable extension development. 
Together, these components lay the groundwork for a sustainable and interoperable data preparation ecosystem that can evolve alongside increasingly complex data-centric AI workflows.

Looking forward, we aim to expand the \textsc{DataFlow}-Ecosystem along multiple modality axes, including \textsc{DataFlow-Table}, \textsc{DataFlow-Graph}, and \textsc{DataFlow-Multimodal}, to support richer data types and workflows.  
We also plan to develop domain-oriented variants, such as \textsc{DataFlow-AI4S} and \textsc{DataFlow-Industry}, tailored for large-scale production environments. 
These extensions will broaden the applicability of \textsc{DataFlow} and strengthen its role as a foundational substrate—and a common protocol—for future research, engineering practice, and community-driven innovation in LLM data preparation.


\clearpage

\bibliographystyle{plainnat}
\bibliography{main}

@article{penedo2024fineweb,
  title={The fineweb datasets: Decanting the web for the finest text data at scale},
  author={Penedo, Guilherme and Kydl{\'\i}{\v{c}}ek, Hynek and Lozhkov, Anton and Mitchell, Margaret and Raffel, Colin A and Von Werra, Leandro and Wolf, Thomas and others},
  journal={Advances in Neural Information Processing Systems},
  volume={37},
  pages={30811--30849},
  year={2024}
}

@article{wettig2024qurating,
  title={Qurating: Selecting high-quality data for training language models},
  author={Wettig, Alexander and Gupta, Aatmik and Malik, Saumya and Chen, Danqi},
  journal={arXiv preprint arXiv:2402.09739},
  year={2024}
}

@misc{jin2020diseasedoespatienthave,
      title={What Disease does this Patient Have? A Large-scale Open Domain Question Answering Dataset from Medical Exams}, 
      author={Di Jin and Eileen Pan and Nassim Oufattole and Wei-Hung Weng and Hanyi Fang and Peter Szolovits},
      year={2020},
      eprint={2009.13081},
      archivePrefix={arXiv},
      primaryClass={cs.CL},
      url={https://arxiv.org/abs/2009.13081}, 
}

@misc{xiong2024benchmarkingretrievalaugmentedgenerationmedicine,
      title={Benchmarking Retrieval-Augmented Generation for Medicine}, 
      author={Guangzhi Xiong and Qiao Jin and Zhiyong Lu and Aidong Zhang},
      year={2024},
      eprint={2402.13178},
      archivePrefix={arXiv},
      primaryClass={cs.CL},
      url={https://arxiv.org/abs/2402.13178}, 
}

@misc{chen2023meditron70bscalingmedicalpretraining,
      title={MEDITRON-70B: Scaling Medical Pretraining for Large Language Models}, 
      author={Zeming Chen and Alejandro Hernández Cano and Angelika Romanou and Antoine Bonnet and Kyle Matoba and Francesco Salvi and Matteo Pagliardini and Simin Fan and Andreas Köpf and Amirkeivan Mohtashami and Alexandre Sallinen and Alireza Sakhaeirad and Vinitra Swamy and Igor Krawczuk and Deniz Bayazit and Axel Marmet and Syrielle Montariol and Mary-Anne Hartley and Martin Jaggi and Antoine Bosselut},
      year={2023},
      eprint={2311.16079},
      archivePrefix={arXiv},
      primaryClass={cs.CL},
      url={https://arxiv.org/abs/2311.16079}, 
}

@inproceedings{jin2019pubmedqa,
  title={Pubmedqa: A dataset for biomedical research question answering},
  author={Jin, Qiao and Dhingra, Bhuwan and Liu, Zhengping and Cohen, William and Lu, Xinghua},
  booktitle={Proceedings of the 2019 conference on empirical methods in natural language processing and the 9th international joint conference on natural language processing (EMNLP-IJCNLP)},
  pages={2567--2577},
  year={2019}
}

@inproceedings{kotonya-toni-2020-explainable-automated,
    title = "Explainable Automated Fact-Checking for Public Health Claims",
    author = "Kotonya, Neema  and
      Toni, Francesca",
    editor = "Webber, Bonnie  and
      Cohn, Trevor  and
      He, Yulan  and
      Liu, Yang",
    booktitle = "Proceedings of the 2020 Conference on Empirical Methods in Natural Language Processing (EMNLP)",
    month = nov,
    year = "2020",
    address = "Online",
    publisher = "Association for Computational Linguistics",
    url = "https://aclanthology.org/2020.emnlp-main.623/",
    doi = "10.18653/v1/2020.emnlp-main.623",
    pages = "7740--7754"
}

@misc{mohr2022covertcorpusfactcheckedbiomedical,
      title={CoVERT: A Corpus of Fact-checked Biomedical COVID-19 Tweets}, 
      author={Isabelle Mohr and Amelie Wührl and Roman Klinger},
      year={2022},
      eprint={2204.12164},
      archivePrefix={arXiv},
      primaryClass={cs.CL},
      url={https://arxiv.org/abs/2204.12164}, 
}

@misc{niu2025mineru25decoupledvisionlanguagemodel,
      title={MinerU2.5: A Decoupled Vision-Language Model for Efficient High-Resolution Document Parsing}, 
      author={Junbo Niu and Zheng Liu and Zhuangcheng Gu and Bin Wang and Linke Ouyang and Zhiyuan Zhao and Tao Chu and Tianyao He and Fan Wu and Qintong Zhang and Zhenjiang Jin and Guang Liang and Rui Zhang and Wenzheng Zhang and Yuan Qu and Zhifei Ren and Yuefeng Sun and Yuanhong Zheng and Dongsheng Ma and Zirui Tang and Boyu Niu and Ziyang Miao and Hejun Dong and Siyi Qian and Junyuan Zhang and Jingzhou Chen and Fangdong Wang and Xiaomeng Zhao and Liqun Wei and Wei Li and Shasha Wang and Ruiliang Xu and Yuanyuan Cao and Lu Chen and Qianqian Wu and Huaiyu Gu and Lindong Lu and Keming Wang and Dechen Lin and Guanlin Shen and Xuanhe Zhou and Linfeng Zhang and Yuhang Zang and Xiaoyi Dong and Jiaqi Wang and Bo Zhang and Lei Bai and Pei Chu and Weijia Li and Jiang Wu and Lijun Wu and Zhenxiang Li and Guangyu Wang and Zhongying Tu and Chao Xu and Kai Chen and Yu Qiao and Bowen Zhou and Dahua Lin and Wentao Zhang and Conghui He},
      year={2025},
      eprint={2509.22186},
      archivePrefix={arXiv},
      primaryClass={cs.CV},
      url={https://arxiv.org/abs/2509.22186}, 
}

@article{ho2020constructing,
  title={Constructing a multi-hop qa dataset for comprehensive evaluation of reasoning steps},
  author={Ho, Xanh and Nguyen, Anh-Khoa Duong and Sugawara, Saku and Aizawa, Akiko},
  journal={arXiv preprint arXiv:2011.01060},
  year={2020}
}

@article{trivedi2022musique,
  title={MuSiQue: Multihop Questions via Single-hop Question Composition},
  author={Trivedi, Harsh and Balasubramanian, Niranjan and Khot, Tushar and Sabharwal, Ashish},
  journal={Transactions of the Association for Computational Linguistics},
  volume={10},
  pages={539--554},
  year={2022},
  publisher={MIT Press One Broadway, 12th Floor, Cambridge, Massachusetts 02142, USA~…}
}

@inproceedings{yang2018hotpotqa,
  title={HotpotQA: A dataset for diverse, explainable multi-hop question answering},
  author={Yang, Zhilin and Qi, Peng and Zhang, Saizheng and Bengio, Yoshua and Cohen, William and Salakhutdinov, Ruslan and Manning, Christopher D},
  booktitle={Proceedings of the 2018 conference on empirical methods in natural language processing},
  pages={2369--2380},
  year={2018}
}

@inproceedings{press2023measuring,
  title={Measuring and narrowing the compositionality gap in language models},
  author={Press, Ofir and Zhang, Muru and Min, Sewon and Schmidt, Ludwig and Smith, Noah A and Lewis, Mike},
  booktitle={Findings of the Association for Computational Linguistics: EMNLP 2023},
  pages={5687--5711},
  year={2023}
}

@article{wang2022text,
  title={Text Embeddings by Weakly-Supervised Contrastive Pre-training},
  author={Wang, Liang and Yang, Nan and Huang, Xiaolong and Jiao, Binxing and Yang, Linjun and Jiang, Daxin and Majumder, Rangan and Wei, Furu},
  journal={arXiv preprint arXiv:2212.03533},
  year={2022}
}

@article{chen2025learning,
  title={Learning to reason with search for llms via reinforcement learning},
  author={Chen, Mingyang and Sun, Linzhuang and Li, Tianpeng and Sun, Haoze and Zhou, Yijie and Zhu, Chenzheng and Wang, Haofen and Pan, Jeff Z and Zhang, Wen and Chen, Huajun and others},
  journal={arXiv preprint arXiv:2503.19470},
  year={2025}
}

@inproceedings{jin2025flashrag,
  title={Flashrag: A modular toolkit for efficient retrieval-augmented generation research},
  author={Jin, Jiajie and Zhu, Yutao and Dou, Zhicheng and Dong, Guanting and Yang, Xinyu and Zhang, Chenghao and Zhao, Tong and Yang, Zhao and Wen, Ji-Rong},
  booktitle={Companion Proceedings of the ACM on Web Conference 2025},
  pages={737--740},
  year={2025}
}

@article{zhuo2024bigcodebench,
  title={Bigcodebench: Benchmarking code generation with diverse function calls and complex instructions},
  author={Zhuo, Terry Yue and Vu, Minh Chien and Chim, Jenny and Hu, Han and Yu, Wenhao and Widyasari, Ratnadira and Yusuf, Imam Nur Bani and Zhan, Haolan and He, Junda and Paul, Indraneil and others},
  journal={arXiv preprint arXiv:2406.15877},
  year={2024}
}

@article{jain2024livecodebench,
  title={Livecodebench: Holistic and contamination free evaluation of large language models for code},
  author={Jain, Naman and Han, King and Gu, Alex and Li, Wen-Ding and Yan, Fanjia and Zhang, Tianjun and Wang, Sida and Solar-Lezama, Armando and Sen, Koushik and Stoica, Ion},
  journal={arXiv preprint arXiv:2403.07974},
  year={2024}
}

@misc{openr1,
    title = {Open R1: A fully open reproduction of DeepSeek-R1},
    url = {https://github.com/huggingface/open-r1},
    author = {{Hugging Face}},
    month = {January},
    year = {2025}
}

@misc{2025synthetic1,
      title={SYNTHETIC-1: Two Million Collaboratively Generated Reasoning Traces from Deepseek-R1}, 
      author={Justus Mattern and Sami Jaghouar and Manveer Basra and Jannik Straube and Matthew Di Ferrante and Felix Gabriel and Jack Min Ong and Vincent Weisser and Johannes Hagemann},
      year={2025},
      url={https://www.primeintellect.ai/blog/synthetic-1-release}, 
}

@article{shen2025let,
  title={Let's Verify Math Questions Step by Step},
  author={Shen, Chengyu and Wong, Zhen Hao and He, Runming and Liang, Hao and Qiang, Meiyi and Meng, Zimo and Zhao, Zhengyang and Zeng, Bohan and Zhu, Zhengzhou and Cui, Bin and others},
  journal={arXiv preprint arXiv:2505.13903},
  year={2025}
}

@article{hendrycks2020measuring,
  title={Measuring massive multitask language understanding},
  author={Hendrycks, Dan and Burns, Collin and Basart, Steven and Zou, Andy and Mazeika, Mantas and Song, Dawn and Steinhardt, Jacob},
  journal={arXiv preprint arXiv:2009.03300},
  year={2020}
}

@misc{alpaca_eval,
  author = {Xuechen Li and Tianyi Zhang and Yann Dubois and Rohan Taori and Ishaan Gulrajani and Carlos Guestrin and Percy Liang and Tatsunori B. Hashimoto },
  title = {AlpacaEval: An Automatic Evaluator of Instruction-following Models},
  year = {2023},
  month = {5},
  publisher = {GitHub},
  journal = {GitHub repository},
  howpublished = {\url{https://github.com/tatsu-lab/alpaca_eval}}
}

@misc{humaneval,
      title={Evaluating Large Language Models Trained on Code}, 
      author={Mark Chen and Jerry Tworek and Heewoo Jun and Qiming Yuan and Henrique Ponde de Oliveira Pinto and Jared Kaplan and Harri Edwards and Yuri Burda and Nicholas Joseph and Greg Brockman and Alex Ray and Raul Puri and Gretchen Krueger and Michael Petrov and Heidy Khlaaf and Girish Sastry and Pamela Mishkin and Brooke Chan and Scott Gray and Nick Ryder and Mikhail Pavlov and Alethea Power and Lukasz Kaiser and Mohammad Bavarian and Clemens Winter and Philippe Tillet and Felipe Petroski Such and Dave Cummings and Matthias Plappert and Fotios Chantzis and Elizabeth Barnes and Ariel Herbert-Voss and William Hebgen Guss and Alex Nichol and Alex Paino and Nikolas Tezak and Jie Tang and Igor Babuschkin and Suchir Balaji and Shantanu Jain and William Saunders and Christopher Hesse and Andrew N. Carr and Jan Leike and Josh Achiam and Vedant Misra and Evan Morikawa and Alec Radford and Matthew Knight and Miles Brundage and Mira Murati and Katie Mayer and Peter Welinder and Bob McGrew and Dario Amodei and Sam McCandlish and Ilya Sutskever and Wojciech Zaremba},
      year={2021},
      eprint={2107.03374},
      archivePrefix={arXiv},
      primaryClass={cs.LG},
      url={https://arxiv.org/abs/2107.03374}, 
}

@article{li2024crowdsourced,
  title={From Crowdsourced Data to High-Quality Benchmarks: Arena-Hard and BenchBuilder Pipeline},
  author={Li, Tianle and Chiang, Wei-Lin and Frick, Evan and Dunlap, Lisa and Wu, Tianhao and Zhu, Banghua and Gonzalez, Joseph E and Stoica, Ion},
  journal={arXiv preprint arXiv:2406.11939},
  year={2024}
}

@misc{lingcodesft,
      title={Every Sample Matters: Leveraging Mixture-of-Experts and High-Quality Data for Efficient and Accurate Code LLM}, 
      author={Codefuse and Ling Team},
      year={2025},
      eprint={2503.17793},
      archivePrefix={arXiv},
      primaryClass={cs.LG},
      url={https://arxiv.org/abs/2503.17793}, 
}

@article{huang2023c,
  title={C-eval: A multi-level multi-discipline chinese evaluation suite for foundation models},
  author={Huang, Yuzhen and Bai, Yuzhuo and Zhu, Zhihao and Zhang, Junlei and Zhang, Jinghan and Su, Tangjun and Liu, Junteng and Lv, Chuancheng and Zhang, Yikai and Fu, Yao and others},
  journal={Advances in Neural Information Processing Systems},
  volume={36},
  pages={62991--63010},
  year={2023}
}

@misc{codealpaca,
  author = {Sahil Chaudhary},
  title = {Code Alpaca: An Instruction-following LLaMA model for code generation},
  year = {2023},
  publisher = {GitHub},
  journal = {GitHub repository},
  howpublished = {\url{https://github.com/sahil280114/codealpaca}},
}

@article{wei2024selfcodealign,
  title={SelfCodeAlign: Self-Alignment for Code Generation}, 
  author={Yuxiang Wei and Federico Cassano and Jiawei Liu and Yifeng Ding and Naman Jain and Zachary Mueller and Harm de Vries and Leandro von Werra and Arjun Guha and Lingming Zhang},
  year={2024},
  journal={arXiv preprint arXiv:2410.24198}
}

@article{paszke2019pytorch,
  title={Pytorch: An imperative style, high-performance deep learning library},
  author={Paszke, Adam and Gross, Sam and Massa, Francisco and Lerer, Adam and Bradbury, James and Chanan, Gregory and Killeen, Trevor and Lin, Zeming and Gimelshein, Natalia and Antiga, Luca and others},
  journal={Advances in neural information processing systems},
  volume={32},
  year={2019}
}

@article{li2025omnisql,
  title={Omnisql: Synthesizing high-quality text-to-sql data at scale},
  author={Li, Haoyang and Wu, Shang and Zhang, Xiaokang and Huang, Xinmei and Zhang, Jing and Jiang, Fuxin and Wang, Shuai and Zhang, Tieying and Chen, Jianjun and Shi, Rui and others},
  journal={arXiv preprint arXiv:2503.02240},
  year={2025}
}

@article{yu2018spider,
  title={Spider: A large-scale human-labeled dataset for complex and cross-domain semantic parsing and text-to-sql task},
  author={Yu, Tao and Zhang, Rui and Yang, Kai and Yasunaga, Michihiro and Wang, Dongxu and Li, Zifan and Ma, James and Li, Irene and Yao, Qingning and Roman, Shanelle and others},
  journal={arXiv preprint arXiv:1809.08887},
  year={2018}
}

@article{li2024can,
  title={Can llm already serve as a database interface? a big bench for large-scale database grounded text-to-sqls},
  author={Li, Jinyang and Hui, Binyuan and Qu, Ge and Yang, Jiaxi and Li, Binhua and Li, Bowen and Wang, Bailin and Qin, Bowen and Geng, Ruiying and Huo, Nan and others},
  journal={Advances in Neural Information Processing Systems},
  volume={36},
  year={2024}
}

@article{lee2022ehrsql,
  title={Ehrsql: A practical text-to-sql benchmark for electronic health records},
  author={Lee, Gyubok and Hwang, Hyeonji and Bae, Seongsu and Kwon, Yeonsu and Shin, Woncheol and Yang, Seongjun and Seo, Minjoon and Kim, Jong-Yeup and Choi, Edward},
  journal={Advances in Neural Information Processing Systems},
  volume={35},
  pages={15589--15601},
  year={2022}
}

@article{gan2021exploring,
  title={Exploring underexplored limitations of cross-domain text-to-SQL generalization},
  author={Gan, Yujian and Chen, Xinyun and Purver, Matthew},
  journal={arXiv preprint arXiv:2109.05157},
  year={2021}
}

@article{gan2021towards,
  title={Towards robustness of text-to-SQL models against synonym substitution},
  author={Gan, Yujian and Chen, Xinyun and Huang, Qiuping and Purver, Matthew and Woodward, John R and Xie, Jinxia and Huang, Pengsheng},
  journal={arXiv preprint arXiv:2106.01065},
  year={2021}
}

@article{deng2020structure,
  title={Structure-grounded pretraining for text-to-SQL},
  author={Deng, Xiang and Awadallah, Ahmed Hassan and Meek, Christopher and Polozov, Oleksandr and Sun, Huan and Richardson, Matthew},
  journal={arXiv preprint arXiv:2010.12773},
  year={2020}
}

@article{grattafiori2024llama,
  title={The llama 3 herd of models},
  author={Grattafiori, Aaron and Dubey, Abhimanyu and Jauhri, Abhinav and Pandey, Abhinav and Kadian, Abhishek and Al-Dahle, Ahmad and Letman, Aiesha and Mathur, Akhil and Schelten, Alan and Vaughan, Alex and others},
  journal={arXiv preprint arXiv:2407.21783},
  year={2024}
}

@article{hui2024qwen2,
  title={Qwen2. 5-coder technical report},
  author={Hui, Binyuan and Yang, Jian and Cui, Zeyu and Yang, Jiaxi and Liu, Dayiheng and Zhang, Lei and Liu, Tianyu and Zhang, Jiajun and Yu, Bowen and Lu, Keming and others},
  journal={arXiv preprint arXiv:2409.12186},
  year={2024}
}

@inproceedings{chen2024data,
  title={Data-juicer: A one-stop data processing system for large language models},
  author={Chen, Daoyuan and Huang, Yilun and Ma, Zhijian and Chen, Hesen and Pan, Xuchen and Ge, Ce and Gao, Dawei and Xie, Yuexiang and Liu, Zhaoyang and Gao, Jinyang and others},
  booktitle={Companion of the 2024 International Conference on Management of Data},
  pages={120--134},
  year={2024}
}

@misc{nvidia2024nemo,
  title        = {Curating Custom Datasets for {LLM} Training with {NVIDIA} NeMo Curator},
  author       = {{NVIDIA Corporation}},
  year         = {2024},
  howpublished = {\url{https://developer.nvidia.com/blog/curating-custom-datasets-for-llm-training-with-nvidia-nemo-curator/}},
  note         = {Accessed: 2025-11-28}
}

@article{du2023mods,
  title={Mods: Model-oriented data selection for instruction tuning},
  author={Du, Qianlong and Zong, Chengqing and Zhang, Jiajun},
  journal={arXiv preprint arXiv:2311.15653},
  year={2023}
}

@article{chen2023alpagasus,
  title={Alpagasus: Training a better alpaca with fewer data},
  author={Chen, Lichang and Li, Shiyang and Yan, Jun and Wang, Hai and Gunaratna, Kalpa and Yadav, Vikas and Tang, Zheng and Srinivasan, Vijay and Zhou, Tianyi and Huang, Heng and others},
  journal={arXiv preprint arXiv:2307.08701},
  year={2023}
}

@article{bai2024survey,
  title={A Survey of Multimodal Large Language Model from A Data-centric Perspective},
  author={Bai, Tianyi and Liang, Hao and Wan, Binwang and Yang, Ling and Li, Bozhou and Wang, Yifan and Cui, Bin and He, Conghui and Yuan, Binhang and Zhang, Wentao},
  journal={arXiv preprint arXiv:2405.16640},
  year={2024}
}

@article{llama,
  title={Llama: Open and efficient foundation language models},
  author={Touvron, Hugo and Lavril, Thibaut and Izacard, Gautier and Martinet, Xavier and Lachaux, Marie-Anne and Lacroix, Timoth{\'e}e and Rozi{\`e}re, Baptiste and Goyal, Naman and Hambro, Eric and Azhar, Faisal and others},
  journal={arXiv preprint arXiv:2302.13971},
  year={2023}
}

@misc{llama3repo,
  author = {meta-llama},
  title = {{Introducing Meta Llama 3: The most capable openly available LLM to date}},
  year = {2024},
  url = {https://ai.meta.com/blog/meta-llama-3/},
  note = {Accessed: 2024-05-02}
}

@article{gao2020pile,
  title={The pile: An 800gb dataset of diverse text for language modeling},
  author={Gao, Leo and Biderman, Stella and Black, Sid and Golding, Laurence and Hoppe, Travis and Foster, Charles and Phang, Jason and He, Horace and Thite, Anish and Nabeshima, Noa and others},
  journal={arXiv preprint arXiv:2101.00027},
  year={2020}
}

@article{li2024datacomp,
  title={Datacomp-lm: In search of the next generation of training sets for language models},
  author={Li, Jeffrey and Fang, Alex and Smyrnis, Georgios and Ivgi, Maor and Jordan, Matt and Gadre, Samir Yitzhak and Bansal, Hritik and Guha, Etash and Keh, Sedrick Scott and Arora, Kushal and others},
  journal={Advances in Neural Information Processing Systems},
  volume={37},
  pages={14200--14282},
  year={2024}
}

@article{achiam2023gpt,
  title={Gpt-4 technical report},
  author={Achiam, Josh and Adler, Steven and Agarwal, Sandhini and Ahmad, Lama and Akkaya, Ilge and Aleman, Florencia Leoni and Almeida, Diogo and Altenschmidt, Janko and Altman, Sam and Anadkat, Shyamal and others},
  journal={arXiv preprint arXiv:2303.08774},
  year={2023}
}

@article{hoffmann2022training,
  title={Training compute-optimal large language models},
  author={Hoffmann, Jordan and Borgeaud, Sebastian and Mensch, Arthur and Buchatskaya, Elena and Cai, Trevor and Rutherford, Eliza and Casas, Diego de Las and Hendricks, Lisa Anne and Welbl, Johannes and Clark, Aidan and others},
  journal={arXiv preprint arXiv:2203.15556},
  year={2022}
}

@article{sorscher2022beyond,
  title={Beyond neural scaling laws: beating power law scaling via data pruning},
  author={Sorscher, Ben and Geirhos, Robert and Shekhar, Shashank and Ganguli, Surya and Morcos, Ari},
  journal={Advances in Neural Information Processing Systems},
  volume={35},
  pages={19523--19536},
  year={2022}
}

@article{yang2025qwen3,
  title={Qwen3 technical report},
  author={Yang, An and Li, Anfeng and Yang, Baosong and Zhang, Beichen and Hui, Binyuan and Zheng, Bo and Yu, Bowen and Gao, Chang and Huang, Chengen and Lv, Chenxu and others},
  journal={arXiv preprint arXiv:2505.09388},
  year={2025}
}

@article{wang2025ultra,
  title={Ultra-fineweb: Efficient data filtering and verification for high-quality llm training data},
  author={Wang, Yudong and Fu, Zixuan and Cai, Jie and Tang, Peijun and Lyu, Hongya and Fang, Yewei and Zheng, Zhi and Zhou, Jie and Zeng, Guoyang and Xiao, Chaojun and others},
  journal={arXiv preprint arXiv:2505.05427},
  year={2025}
}

@inproceedings{wang2023self,
  title={Self-instruct: Aligning language models with self-generated instructions},
  author={Wang, Yizhong and Kordi, Yeganeh and Mishra, Swaroop and Liu, Alisa and Smith, Noah A and Khashabi, Daniel and Hajishirzi, Hannaneh},
  booktitle={Proceedings of the 61st annual meeting of the association for computational linguistics (volume 1: long papers)},
  pages={13484--13508},
  year={2023}
}

@article{yu2025cot,
  title={Cot-self-instruct: Building high-quality synthetic prompts for reasoning and non-reasoning tasks},
  author={Yu, Ping and Lanchantin, Jack and Wang, Tianlu and Yuan, Weizhe and Golovneva, Olga and Kulikov, Ilia and Sukhbaatar, Sainbayar and Weston, Jason and Xu, Jing},
  journal={arXiv preprint arXiv:2507.23751},
  year={2025}
}

@inproceedings{ostendorff2024llm,
  title={LLM-datasets: An open framework for pretraining datasets of large language models},
  author={Ostendorff, Malte and Suarez, Pedro Ortiz and Lage, Lucas Fonseca and Rehm, Georg},
  booktitle={First conference on language modeling},
  year={2024}
}

@article{zhang2023instruction,
  title={Instruction tuning for large language models: A survey},
  author={Zhang, Shengyu and Dong, Linfeng and Li, Xiaoya and Zhang, Sen and Sun, Xiaofei and Wang, Shuhe and Li, Jiwei and Hu, Runyi and Zhang, Tianwei and Wang, Guoyin and others},
  journal={ACM Computing Surveys},
  year={2023},
  publisher={ACM New York, NY}
}

@article{wei2022chain,
  title={Chain-of-thought prompting elicits reasoning in large language models},
  author={Wei, Jason and Wang, Xuezhi and Schuurmans, Dale and Bosma, Maarten and Xia, Fei and Chi, Ed and Le, Quoc V and Zhou, Denny and others},
  journal={Advances in neural information processing systems},
  volume={35},
  pages={24824--24837},
  year={2022}
}

@article{li2025infinity,
  title={Infinity Instruct: Scaling Instruction Selection and Synthesis to Enhance Language Models},
  author={Li, Jijie and Du, Li and Zhao, Hanyu and Zhang, Bo-wen and Wang, Liangdong and Gao, Boyan and Liu, Guang and Lin, Yonghua},
  journal={arXiv preprint arXiv:2506.11116},
  year={2025}
}

@article{gu2024cruxeval,
  title={Cruxeval: A benchmark for code reasoning, understanding and execution},
  author={Gu, Alex and Rozi{\`e}re, Baptiste and Leather, Hugh and Solar-Lezama, Armando and Synnaeve, Gabriel and Wang, Sida I},
  journal={arXiv preprint arXiv:2401.03065},
  year={2024}
}

@article{cobbe2021training,
  title={Training verifiers to solve math word problems},
  author={Cobbe, Karl and Kosaraju, Vineet and Bavarian, Mohammad and Chen, Mark and Jun, Heewoo and Kaiser, Lukasz and Plappert, Matthias and Tworek, Jerry and Hilton, Jacob and Nakano, Reiichiro and others},
  journal={arXiv preprint arXiv:2110.14168},
  year={2021}
}

@article{hendrycks2021measuring,
  title={Measuring mathematical problem solving with the math dataset},
  author={Hendrycks, Dan and Burns, Collin and Kadavath, Saurav and Arora, Akul and Basart, Steven and Tang, Eric and Song, Dawn and Steinhardt, Jacob},
  journal={arXiv preprint arXiv:2103.03874},
  year={2021}
}

@software{langgraph2024,
  author = {LangChain AI},
  title = {LangGraph},
  year = {2024},
  publisher = {GitHub},
  journal = {GitHub repository},
  howpublished = {\url{https://github.com/langchain-ai/langgraph}},
}

@misc{gamma1995design,
  title={Design patterns: elements of reusable object-oriented software},
  author={Gamma, Erich},
  year={1995},
  publisher={Addison-Wesley}
}

@inproceedings{kwon2023efficient_vllm,
  title={Efficient memory management for large language model serving with pagedattention},
  author={Kwon, Woosuk and Li, Zhuohan and Zhuang, Siyuan and Sheng, Ying and Zheng, Lianmin and Yu, Cody Hao and Gonzalez, Joseph and Zhang, Hao and Stoica, Ion},
  booktitle={Proceedings of the 29th symposium on operating systems principles},
  pages={611--626},
  year={2023}
}

@article{zheng2024sglang,
  title={Sglang: Efficient execution of structured language model programs},
  author={Zheng, Lianmin and Yin, Liangsheng and Xie, Zhiqiang and Sun, Chuyue Livia and Huang, Jeff and Yu, Cody Hao and Cao, Shiyi and Kozyrakis, Christos and Stoica, Ion and Gonzalez, Joseph E and others},
  journal={Advances in neural information processing systems},
  volume={37},
  pages={62557--62583},
  year={2024}
}

@article{team2024gemini,
  title={Gemini: A family of highly capable multimodal models, 2024},
  author={Team, Gemini and Anil, R and Borgeaud, S and Wu, Y and Alayrac, JB and Yu, J and Soricut, R and Schalkwyk, J and Dai, AM and Hauth, A and others},
  journal={arXiv preprint arXiv:2312.11805},
  volume={10},
  year={2024}
}

@article{cai2025text2sql,
  title={Text2SQL-Flow: A Robust SQL-Aware Data Augmentation Framework for Text-to-SQL},
  author={Cai, Qifeng and Liang, Hao and Xu, Chang and Xie, Tao and Zhang, Wentao and Cui, Bin},
  journal={arXiv preprint arXiv:2511.10192},
  year={2025}
}

@article{Spark,
author = {Zaharia, Matei and Xin, Reynold S. and Wendell, Patrick and Das, Tathagata and Armbrust, Michael and Dave, Ankur and Meng, Xiangrui and Rosen, Josh and Venkataraman, Shivaram and Franklin, Michael J. and Ghodsi, Ali and Gonzalez, Joseph and Shenker, Scott and Stoica, Ion},
title = {Apache Spark: a unified engine for big data processing},
year = {2016},
issue_date = {November 2016},
publisher = {Association for Computing Machinery},
address = {New York, NY, USA},
volume = {59},
number = {11},
issn = {0001-0782},
url = {https://doi.org/10.1145/2934664},
doi = {10.1145/2934664},
journal = {Commun. ACM},
month = oct,
pages = {56–65},
numpages = {10}
}

@inproceedings{rocklin2015dask,
  title={Dask: Parallel computation with blocked algorithms and task scheduling.},
  author={Rocklin, Matthew and others},
  booktitle={SciPy},
  pages={126--132},
  year={2015}
}

@inproceedings{ghemawat2003google,
  title={The Google file system},
  author={Ghemawat, Sanjay and Gobioff, Howard and Leung, Shun-Tak},
  booktitle={Proceedings of the nineteenth ACM symposium on Operating systems principles},
  pages={29--43},
  year={2003}
}

@article{dean2008mapreduce,
  title={MapReduce: simplified data processing on large clusters},
  author={Dean, Jeffrey and Ghemawat, Sanjay},
  journal={Communications of the ACM},
  volume={51},
  number={1},
  pages={107--113},
  year={2008},
  publisher={ACM New York, NY, USA}
}

@book{white2012hadoop,
  title={Hadoop: The definitive guide},
  author={White, Tom},
  year={2012},
  publisher={" O'Reilly Media, Inc."}
}

@article{zheng2024llamafactory,
  title={Llamafactory: Unified efficient fine-tuning of 100+ language models},
  author={Zheng, Yaowei and Zhang, Richong and Zhang, Junhao and Ye, Yanhan and Luo, Zheyan and Feng, Zhangchi and Ma, Yongqiang},
  journal={arXiv preprint arXiv:2403.13372},
  year={2024}
}

@article{sheng2024hybridflow,
  title   = {HybridFlow: A Flexible and Efficient RLHF Framework},
  author  = {Guangming Sheng and Chi Zhang and Zilingfeng Ye and Xibin Wu and Wang Zhang and Ru Zhang and Yanghua Peng and Haibin Lin and Chuan Wu},
  year    = {2024},
  journal = {arXiv preprint arXiv: 2409.19256}
}

\clearpage

\beginappendix

\section{Author Contributions}

\newcommand{\ProjectLeader}{\textcolor{red!70!black}{\textit{Project Leader}}}
\newcommand{\ProjectFounder}{\textcolor{blue!70!black}{\textit{Project Founder}}}
\newcommand{\CoreContributor}{\textcolor{green!50!black}{\textit{Core Contributor}}}
\newcommand{\Contributor}{\textcolor{green!80!black}{\textit{Contributor}}}
\newcommand{\ProjectSupervisor}{\textcolor{purple!70!black}{\textit{Project Supervisor}}}
\newcommand{\CorrespondingAuthor}{\textcolor{orange!80!black}{\textit{Corresponding Author}}}


\begin{itemize}
    \item Hao Liang: \ProjectLeader, \ProjectFounder; algorithm lead and manuscript writing.
    \item Xiaochen Ma: \ProjectLeader, \ProjectFounder; system lead and manuscript writing.
    \item Zhou Liu: \ProjectLeader, \ProjectFounder; \textsc{DataFlow-Agent} lead and manuscript writing.

    \item Zhen Hao Wong: \CoreContributor, \ProjectFounder; designs and develops reasoning pipelines, AI4S pipelines, and AgenticRAG pipelines.
    \item Zhengyang Zhao: \CoreContributor, \ProjectFounder; designs and conducts experiments for the Text Pipeline.
    \item Zimo Meng: \CoreContributor, \ProjectFounder; system development and support for the design and experiments of the Text Pipeline.
    \item Runming He: \CoreContributor, \ProjectFounder; develops reasoning pipelines and conducts math reasoning experiments.
    \item Chengyu Shen: \CoreContributor, \ProjectFounder; \textsc{DataFlow} evaluation pipelines and reasoning pipelines.

    \item Qifeng Cai: \CoreContributor; designs and conducts experiments for the Text-to-SQL Pipeline.
    \item Zhaoyang Han: \CoreContributor; designs and conducts experiments for the Knowledge Cleaning Pipeline.
    \item Meiyi Qiang: \CoreContributor; scientific visualization, publicity leadership, and testing.
    \item Yalin Feng: \CoreContributor; designs \textsc{DataFlow} evaluation and PDF2Model pipelines.
    \item Tianyi Bai: \CoreContributor; designs and conducts experiments for the Code Pipeline.

    \item Zewei Pan: \Contributor; designs and conducts the operator-writing workflow and experiments for \textsc{DataFlow-Agent}.
    \item Ziyi Guo: \Contributor; designs and conducts the operator-reuse workflow for \textsc{DataFlow-Agent}.
    \item Yizhen Jiang: \Contributor; supports the design and experiments for the Code Pipeline.
    \item Jingwen Deng: \Contributor; develops the VQA extraction pipeline and operators.
    \item Qijie You: \Contributor; develops the AgenticRAG pipeline and conducts experiments.
    \item Peichao Lai: \Contributor; develops the frontend of \textsc{DataFlow-WebUI}.
    \item Tianyu Guo: \Contributor; develops audio-to-text operators.
    \item Chi Hsu Tsai: \Contributor; fixes bugs and applies \textsc{DataFlow} to achieve first place in the BAAI LIC Challenge.
    \item Hengyi Feng: \Contributor; \textsc{DataFlow} testing.
    \item Rui Hu: \Contributor; conducts \textsc{DataFlow-Instruct-10K} experiments.
    \item Wenkai Yu: \Contributor; implements several operators.
    \item Junbo Niu: \Contributor; supports the integration of MinerU into the Knowledge Cleaning Pipeline.
    \item Bohan Zeng: \Contributor; supports framework design and provides serving for text-to-image components.
    \item Ruichuan An: \Contributor; supports framework design and provides VQA-related component design.
    \item Lu Ma: \Contributor; implements several operators.
    \item Jihao Huang: \Contributor; integrates LightRAG serving.
    \item Yaowei Zheng: \Contributor; integration of \textsc{DataFlow} data with LLaMA-Factory.
    \item Conghui He: \ProjectSupervisor; project supervision and integration of MinerU with \textsc{DataFlow}.
    \item Linpeng Tang: \ProjectSupervisor; project supervision.
    \item Bin Cui: \ProjectSupervisor; project supervision.
    \item Weinan E:  \ProjectSupervisor; project supervision.
    \item Wentao Zhang: \CorrespondingAuthor, \ProjectSupervisor; manuscript writing and project supervision.
\end{itemize}

\end{document}